\documentclass[lettersize,journal]{IEEEtran}

\usepackage{amsmath,amssymb,amsfonts}
\usepackage{algorithmic}
\usepackage{algorithm}
\usepackage{array}
\usepackage[caption=false,font=normalsize,labelfont=sf,textfont=sf]{subfig}
\usepackage{textcomp}
\usepackage{stfloats}
\usepackage{url}
\usepackage{verbatim}
\usepackage{graphicx}
\usepackage{cite}

\usepackage{xcolor}
\usepackage{booktabs}
\usepackage{makecell}
\usepackage{hyperref}
\hypersetup{hidelinks=true}

\definecolor{orange}{rgb}{1,0.5,0}

\def\BibTeX{{\rm B\kern-.05em{\sc i\kern-.025em b}\kern-.08em
    T\kern-.1667em\lower.7ex\hbox{E}\kern-.125emX}}

\newif\ifanonymous
\anonymousfalse

\begin{document}
\title{TAMI: Temporally Aligned, Missingness-Aware, and Interpretable Multimodal Fusion for Mental Health Assessment in Older Adults with Mild Cognitive Impairment}
\author{Merna Bibars, Bolaji Omofojoye, Allan I. Levey, Rachel Hershenberg, Gari D. Clifford,~\IEEEmembership{Fellow,~IEEE}, and Hyeokhyen Kwon
\thanks{Hyeokhyen Kwon, Allan I. Levey, Rachel Hershenberg, Bolaji Omofojoye, and Gari Clifford are partially funded by the National Institute on Deafness and Other Communication Disorders (grant \# 1R21DC021029-01A1). 
The Cognitive Empowerment Program is supported by a generous investment from the James M. Cox Foundation and Cox Enterprises, Inc.
The authors thank the participants of the Charlie and Harriet Shaffer Cognitive Empowerment Program at Emory University. }
\thanks{Merna Bibars is with the Wallace H. Coulter Department of Biomedical Engineering, Georgia Institute of Technology and Emory University, Atlanta, GA, USA, and the Department of Systems and Biomedical Engineering, Cairo University, Giza, Egypt.}
\thanks{Bolaji Omofojoye is with the Department of Biomedical Informatics, Emory University, Atlanta, GA, USA.}
\thanks{Allan I. Levey is with the Department of Neurology, Emory University, Atlanta, GA, USA.}
\thanks{Rachel Hershenberg is with the 
Department of Psychiatry and Behavioral Sciences, Emory University, Atlanta, GA, USA.}
\thanks{Gari D. Clifford is with the Wallace H. Coulter Department of Biomedical Engineering, 
Georgia Institute of Technology, and the Department of Biomedical Informatics, Emory University, Atlanta, GA, USA.}
\thanks{Hyeokhyen Kwon is with 
the Wallace H. Coulter Department of Biomedical Engineering, Georgia Institute of Technology, and the Department of Biomedical Informatics, Emory University, Atlanta, GA, USA.}
}

\markboth{IEEE TRANSACTIONS ON AFFECTIVE COMPUTING}
{Bibars \MakeLowercase{et al.}: TAMI}

\maketitle

\begin{abstract}
Depression and anxiety in older adults with Mild Cognitive Impairment (MCI) are frequently underdiagnosed due to limited access to care. Multimodal analysis of remote clinical interviews is a scalable screening approach, but existing methods have three limitations. First, they do not correct temporal misalignment across multimodal features extracted at different resolutions, inducing spurious cross-modal associations. Second, remote recordings exhibit uneven modality dropout, but missing values are often zero-filled, making them indistinguishable from valid near-zero measurements. Finally, they do not jointly attribute predictions to modalities, questions, and interview moments, limiting fine-grained clinical interpretation. We propose a Temporally-Aligned, Missingness-Aware, Interpretable (TAMI) multimodal fusion framework. TAMI aligns speech, language, facial, and physiological features within question-answer segments on a shared timeline, encodes modality-level missingness over time, and conditions fusion on question context. In interviews with 49 older adults with MCI, TAMI achieved area under the receiver operating characteristic curve (AUROC) scores of $0.68 \pm 0.04$ (depression) and $0.69 \pm 0.09$ (anxiety). Fine-grained temporal alignment of multimodal features produced the largest performance gain ($\Delta{\geq}0.1$ AUROC). Multi-level interpretability analysis revealed that depression classification relied on eyegaze and open-ended questions, while anxiety classification depended on eyegaze and head pose, with attribution uniformly distributed across questions. Using only responses to the open-ended questions (5.1 min), the depression model achieved an AUROC score of $0.67$, which was not significantly different from using the full interview (19 min) ($p>0.05$). Our findings support designing interview protocols centered on open-ended questions for depression screening in older adults with MCI. 



\end{abstract}


\section{Introduction}
\label{sec:introduction} 


\begin{figure}[t]
    \centering
    \includegraphics[width=\columnwidth]{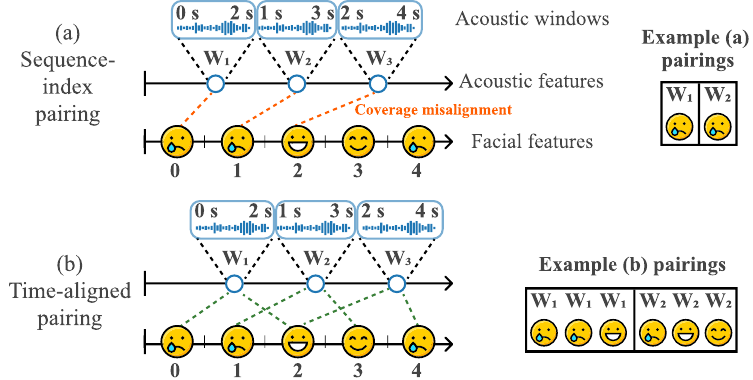}
    \caption{
    Temporal misalignment in multimodal behavioral feature processing in existing work. 
    Conventionally, acoustic features are extracted with overlapping sliding windows (2s window, 1s hop), e.g., $W_1$ (spanning 0--2s), while facial features are frame-level observations sampled at 1 fps. (a) Existing work uses sequence-index pairing to associate feature time series by their position in each series rather than by recording time, linking each acoustic window to the single facial frame at the same index (\textcolor{orange}{dotted orange line}). 
    This produces coverage misalignment in behavioral analysis.
    (b) Our proposed multimodal temporal binning approach assigns features according to their temporal support based on the portion of the behavioral recording that each feature represents. Each acoustic window is associated with the facial frames occurring within the same time interval (\textcolor{teal}{dotted teal lines}). 
    Such temporal alignment of multimodal behavioral features is critical for capturing fine-grained behavioral congruence during remote mental health interviews.
    }
    \label{fig:align_reason}
\end{figure}
\IEEEPARstart{M}{ild} Cognitive Impairment (MCI), a clinical condition that precedes Alzheimer's disease and related dementias, is projected to affect 21 million adults in the United States (U.S.) by 2060~\cite{rajan2021admci}. It is characterized by cognitive decline greater than expected for healthy aging, while preserving daily functioning~\cite{gauthier2006mci}. Depression affects 32\% of adults with MCI~\cite{ismail2017depressionmci, martin2020npsmci}, while anxiety affects 10--50\%, and both are associated with increased risk of progression to dementia~\cite{ma2020neuropsychiatricmci}. Despite their clinical relevance, they remain underdiagnosed~\cite{starkstein2006depressionalzheimers, bassil2011anxietyolderadults}, and older adults in the U.S. face multiple barriers to mental health care~\cite{lavingia2020barriers}. Diagnosis is further complicated by atypical late-life presentations: ``depression without sadness''~\cite{bergua2026dsmdepressionolderadults} and anxiety profiles poorly captured by criteria developed from younger populations~\cite{bryant2013anxietydsm5olderadults}.

\begin{figure*}[t]
    \centering

    \begin{minipage}[t]{0.43\textwidth}
        \centering
        \includegraphics[width=\linewidth]{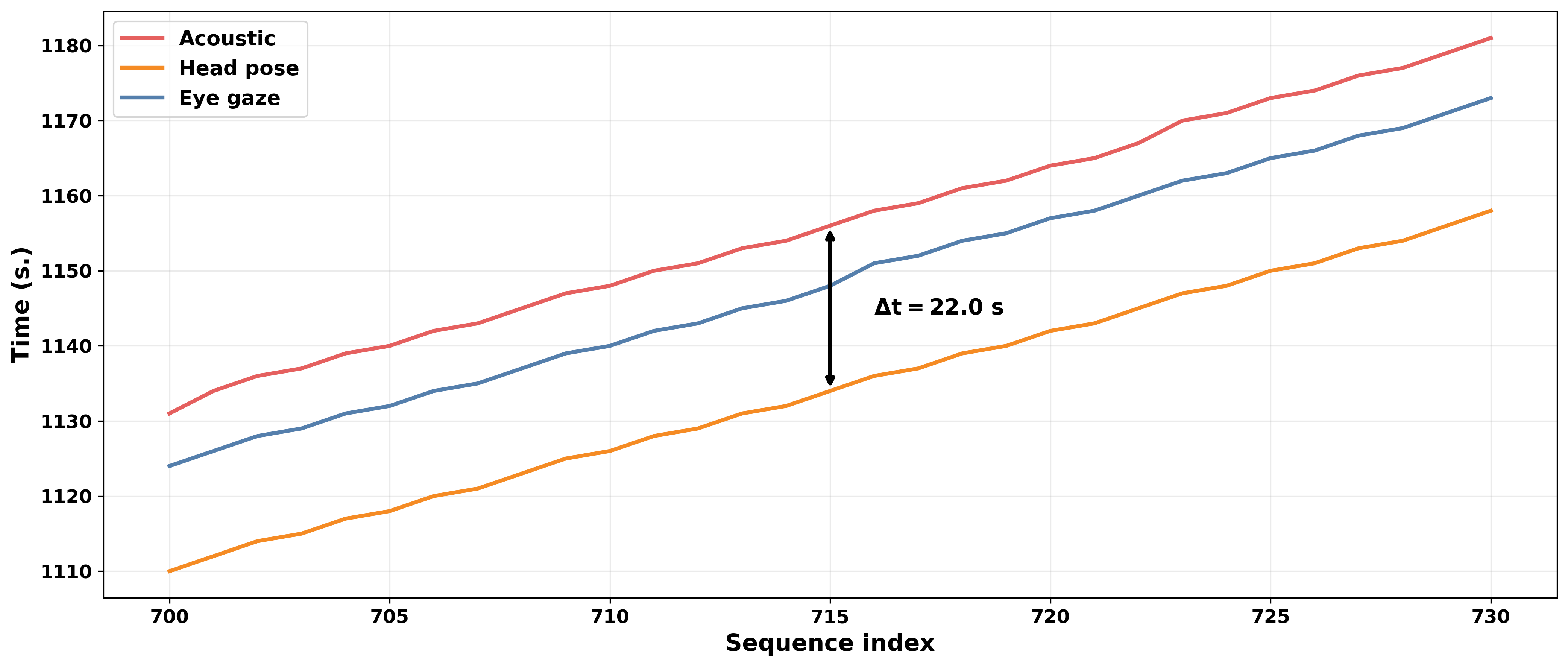}

        \vspace{0.4em}
        \small \textbf{(a)} Temporal displacement
    \end{minipage}
    \hfill
    \begin{minipage}[t]{0.54\textwidth}
        \centering

        \includegraphics[width=0.32\linewidth,height=2.7cm,keepaspectratio]{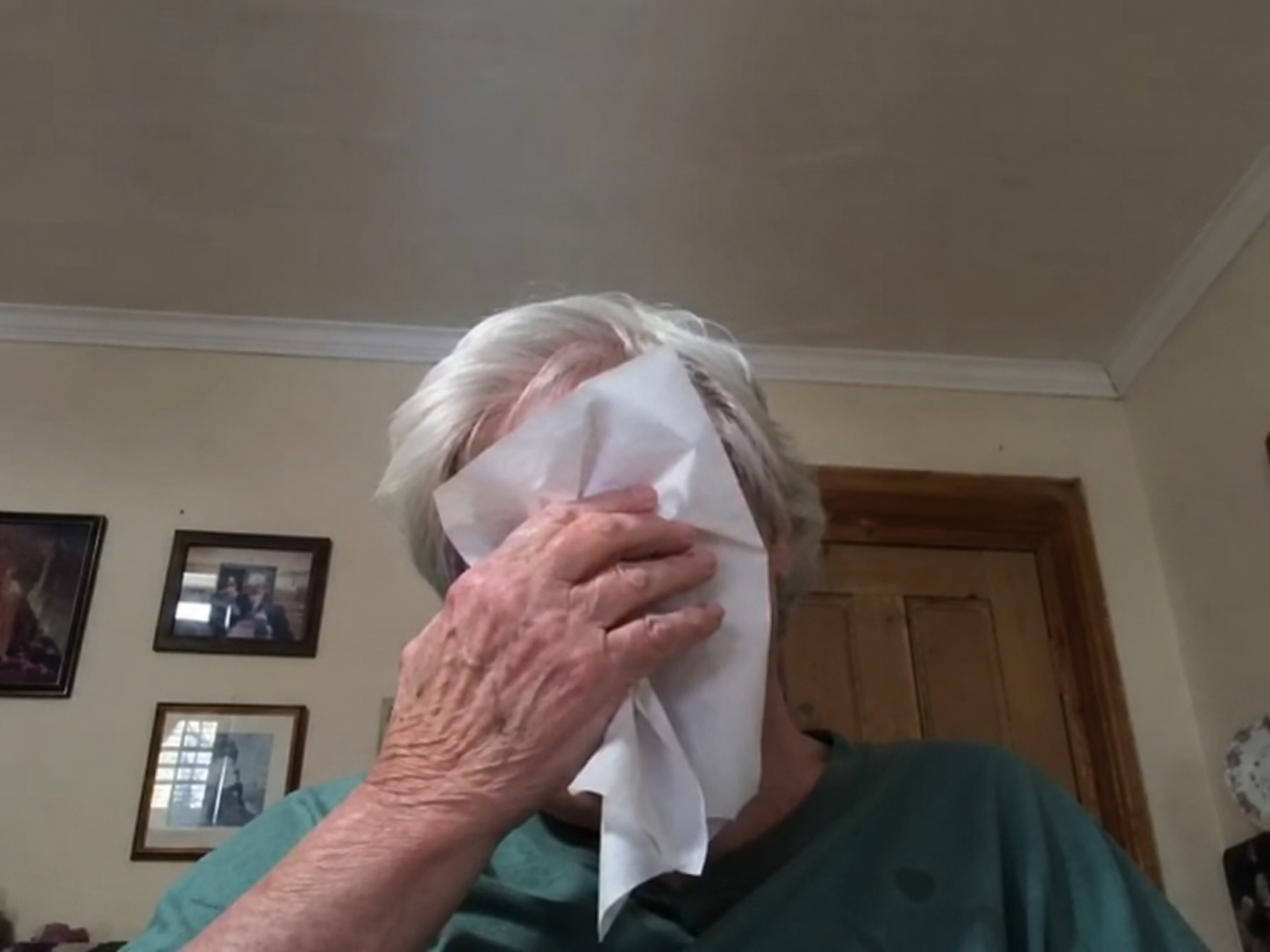}
        \hspace{0.02\linewidth}
        \includegraphics[width=0.20\linewidth,height=2.7cm,keepaspectratio]{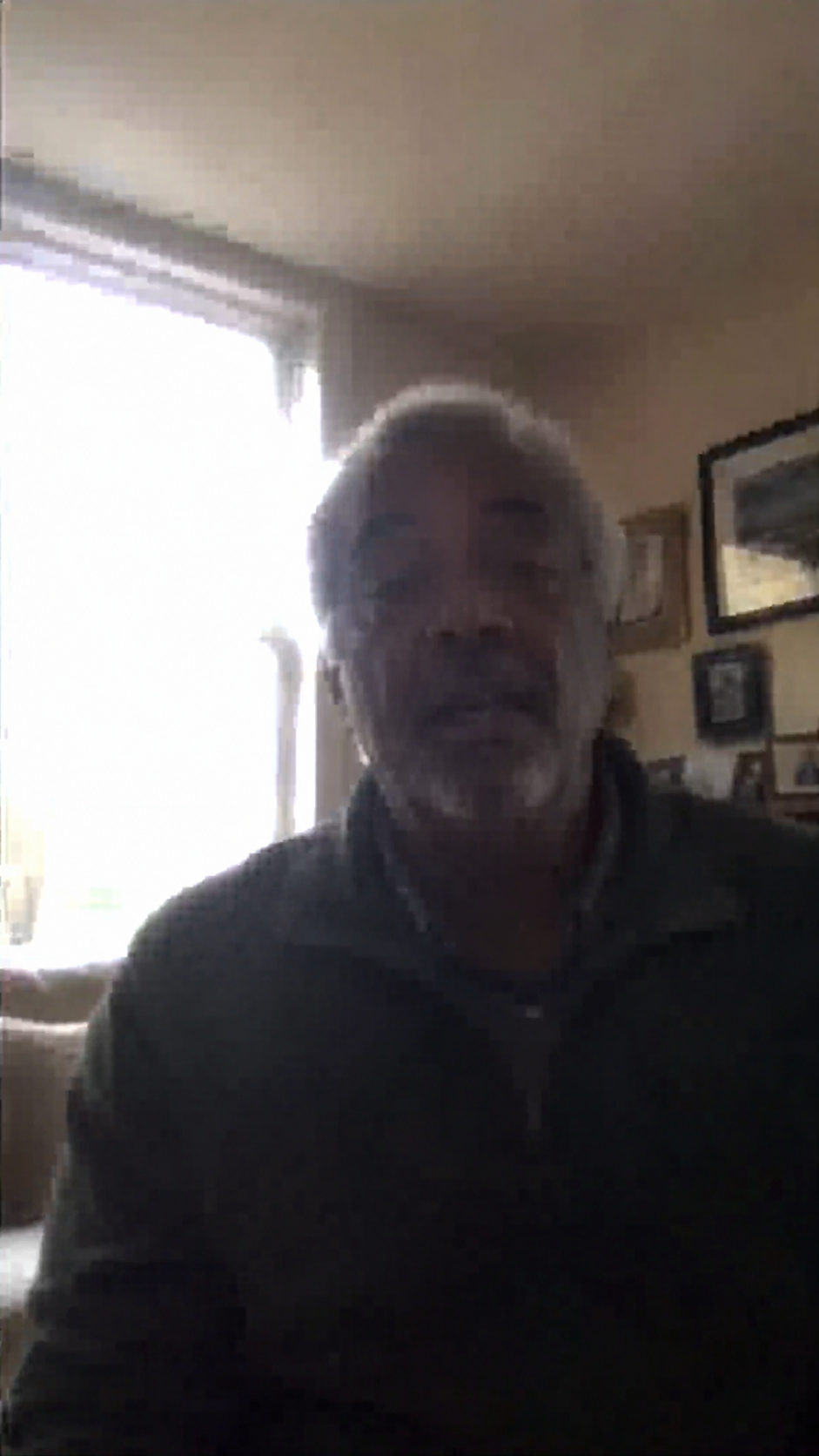}
        \hspace{0.02\linewidth}
        \includegraphics[width=0.32\linewidth,height=2.7cm,keepaspectratio]{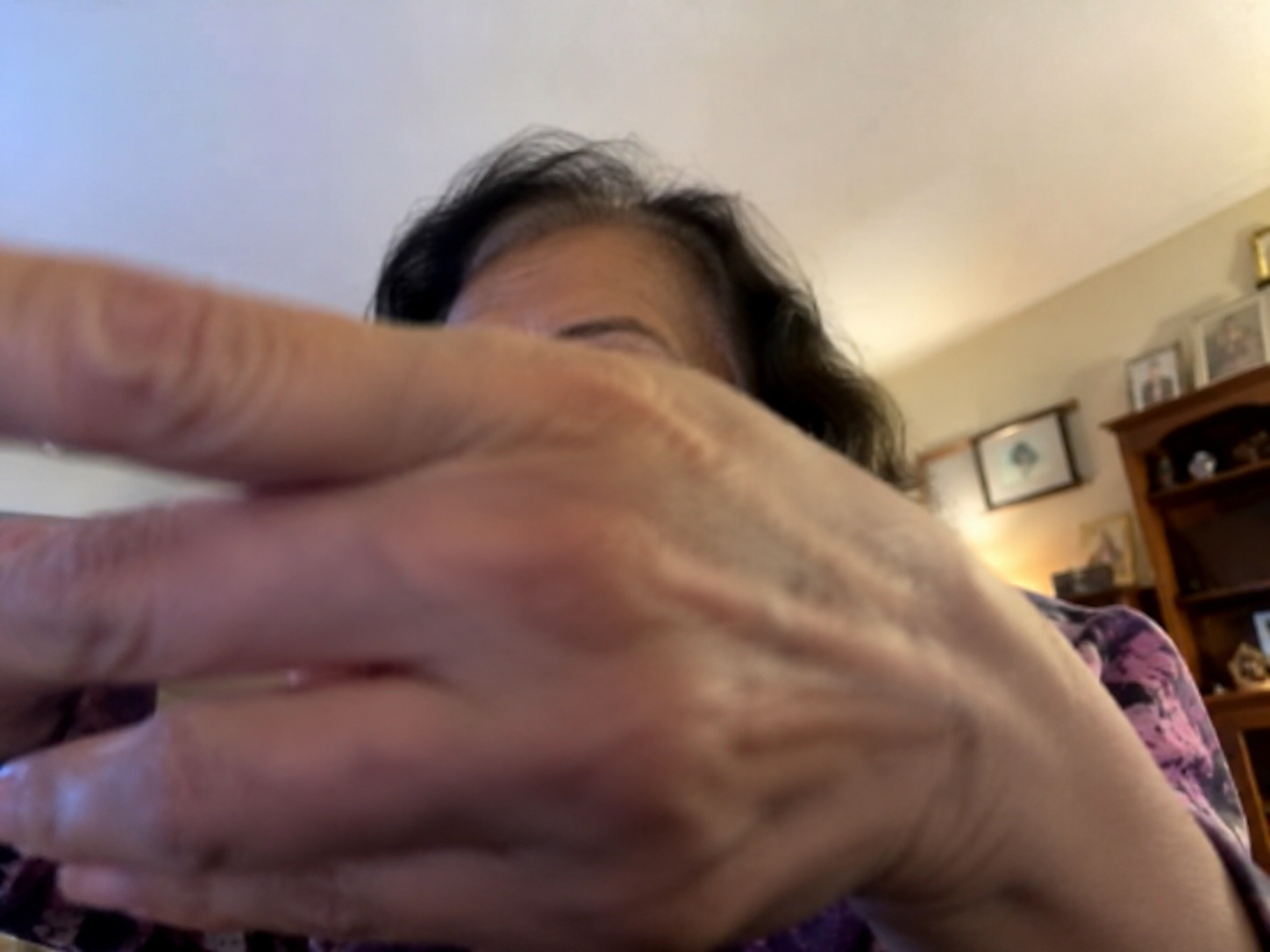}

        \vspace{0.4em}
        \small \textbf{(b)} Visual degradation from occlusion, blur, and overexposure
    \end{minipage}

    \caption{Challenges in multimodal interview analysis. 
    (a) Existing methods associate misaligned behavioral features with temporal displacements of up to 22 seconds (more than 100 times the 200 ms window over which clinicians perceptually bind cross-modal behavioral cues).
    (b) Synthetic illustrations of visual degradation in remote interviews. Privacy constraints prevent the display of participant recordings, so we generate illustrative frames using OpenAI ChatGPT Images 2.0, accessed via ChatGPT~\cite{openai2026chatgptimages2}, to reproduce artifacts observed in the real data, including face occlusion, low resolution with overexposure, and blur. }
    \label{fig:alignment_missingness_examples}
\end{figure*}

Clinical assessment of mental health depends on behaviors that are hard to observe in real time. For example, depression-related changes in speech and eye movements lie in the hundreds-of-milliseconds range~\cite{alpert2001acousticdepression, li2016eyemovementdepression}, too brief to be reliably observed by a clinician during an interview or documented in a clinical report~\cite{alpert2001acousticdepression}. Assessment also relies on affect, where the clinician judges whether a patient's nonverbal behavior is congruent with their verbally stated mood~\cite{voss2024mse}. Depressed and anxious individuals may smile and not show sadness while describing sad experiences~\cite{tremeau2005facialexpressiveness} and anxious events~\cite{harrigan1996facialanxiety}. Reading this congruence requires interpreting modalities jointly and at the moment they co-occur. Human perception binds audiovisual speech into a single event only within a window of 200 ms~\cite{vanwassenhove2007audiovisualspeech}, and behavioral cues separated by intervals longer than 200 ms are unlikely to be perceived as a single event. Therefore, detecting congruence across modalities requires fine-grained and temporally aligned multimodal measurement.


Remote video interviews provide a practical setting for measuring these behavioral signals, given that telehealth use among older adults has increased after the COVID-19 pandemic~\cite{choi2022telehealtholderadults} and that an estimated 140 million Americans have limited access to mental health care~\cite{hrsahealthworkforceshortage}. Recent work has shown that multimodal machine learning (ML) methods can objectively quantify facial, vocal, language, and physiological cues from video interviews~\cite{jiang2024multimodaldigitalbiomarkers,mu2026remotecognitivewellbeing}, providing measures that may support scalable screening of individuals at risk for psychiatric episodes or suicidal behavior, facilitate earlier triage, and complement clinician assessments~\cite{low2020automatedspeechpsychiatric}.

However, prior multimodal ML approaches for remote video interviews remain limited in their ability to analyze multimodal behavioral time series associated with mental health in MCI. Multimodal features are extracted at different temporal resolutions and supports (time intervals over which a feature is computed): window-level features summarize signal content over an interval (e.g., acoustic or physiological features extracted with sliding windows over several seconds), frame-level features are measured at individual timestamps (e.g., facial features extracted at 1 frame per second (fps)), and word-level language features occupy variable temporal spans. Existing approaches temporally associate multimodal features by sequence-index pairing, aligning the features by their sequence position in each extracted feature time series rather than by the timestamps to which they correspond in the original recording~\cite{guo2022topicattentivedepression, jung2024hiquedepression, zhang2025interviewerbiasdepression, fan2024transformermultimodaldepression, tao2024depmstat, gimenogomez2024multimodaldepressionvideo}. 
Consequently, this sequence-index pairing creates temporal coverage misalignment (or displacement) across multimodal behavioral features, in which a multi-second window-level feature (e.g., acoustic) is associated with a single frame-level observation (e.g., facial), despite the window covering multiple observations in time, as shown in \autoref{fig:align_reason}(a).  
\autoref{fig:alignment_missingness_examples}(a) also shows this issue for an example video: across a segment of consecutive indices, the acoustic, head pose, and eyegaze features assigned to the same index (e.g., 715) were recorded up to 22 seconds apart, more than 100 times the 200 ms window within which humans perceptually bind cross-modal behavioral cues~\cite{vanwassenhove2007audiovisualspeech}. This causes the model to fuse features from different time segments of the response, thereby learning misaligned behavioral co-occurrences. In contrast, a temporally aligned representation fuses concurrent behavioral cues, supporting clinically meaningful interpretation across frame-, word-, and window-level behavioral features, as shown in \autoref{fig:align_reason}(b).

Interpretability is necessary for clinicians to trust a model's prediction~\cite{joyce2023explainableai}. It requires attributing a prediction across modalities, questions, and time, so a clinician can identify which behavioral signals, in response to which questions, and at which moments in the interview drove the prediction. Comprehensive semi-structured interviews that combine open-ended questions with multiple screening instruments may require substantial administration time. In our protocol, interviews lasted from 21 to 67 minutes. Therefore, identifying which portions of the interview contain sufficient predictive evidence may inform shorter protocols that reduce clinician time and participant burden. These question- and time-level attributions are meaningful only when the modalities share a common timeline, making fine-grained cross-modal temporal alignment a prerequisite for interpretation.

Remote video recordings collected in real-world settings also exhibit missingness across modalities over time. Participants complete interviews under varying lighting conditions and camera placements so that the video may include blur and occlusion (\autoref{fig:alignment_missingness_examples}(b)), and unstable internet connectivity may cause frames to drop~\cite{cheng2024gracevideocommunication}. Under these conditions, feature extraction may fail at specific timesteps across the modalities~\cite{gimenogomez2024multimodaldepressionvideo, jiang2024multimodaldigitalbiomarkers}.
To handle missingness, previous studies zero-filled missing values~\cite{kumar2025interpretabledepression}, which makes true absence indistinguishable from a valid near-zero measurement. A model must therefore explicitly encode feature missingness.

Furthermore, participant answers are also question-dependent: a brief yes/no answer can carry different clinical meanings depending on the prompt that elicited it. The answer to a question is the co-occurring cross-modal behavior described above, so the question is more informative when it conditions the multimodal sequence at the moment the behavior occurs. Prior work on mental health prediction in the MCI population does not incorporate the question context~\cite{mu2026remotecognitivewellbeing,zhou2022apathydepressionmci,zhou2023speechfacialmci}. Some of the existing depression prediction approaches incorporate question context, but do not apply this conditioning jointly across modalities at a fine-grained temporal resolution. One approach combines the question with temporally pooled text and audio representations for each answer segment~\cite{zhang2025interviewerbiasdepression}. Other approaches condition only a subset of the modalities on the question. For example, Guo \textit{et al.}~\cite{guo2022topicattentivedepression} condition only the text modality on the question in their multimodal approach, while Niu \textit{et al.}~\cite{niu2021hcagdepression} condition audio and text in separate models before fusion. As a result, current approaches do not allow the question to influence the joint cross-modal behavior at the moment it occurs during modeling.

We propose TAMI, a Temporally Aligned, Missingness-Aware, and Interpretable multimodal fusion framework for predicting depression and anxiety in older adults with MCI from remote interviews. TAMI integrates fine-grained cross-modal temporal alignment of features with heterogeneous sampling rates, modality-level missingness over time, and question conditioning.

\section{Related Work}
\label{sec:related_work}


\subsection{Multimodal Mental Health Interview Analysis}

Automated prediction of depression and anxiety from recorded interviews has been studied primarily in general adult populations using datasets such as DAIC-WOZ~\cite{gratch2014daic, devault2014simsensei, ringeval2019avec, gimenogomez2024multimodaldepressionvideo, zhang2025interviewerbiasdepression}. In contrast, interview-based pipelines in older adults with MCI have largely targeted cognitive assessment~\cite{sun2024mcvivit, alsuhaibani2024stam,jiang2024visualexploration}, whereas mental health prediction in this population remains underexplored~\cite{zhou2022apathydepressionmci, zhou2023speechfacialmci,mu2026remotecognitivewellbeing}. Across both populations, multimodal approaches combine facial, acoustic, language, and physiological remote photoplethysmography (rPPG) features through late, early, or mid-fusion~\cite{williamson2016multimodaldepression, drougkas2024multimodalmentalhealth, kumar2025interpretabledepression, jung2024hiquedepression, fan2024transformermultimodaldepression, gimenogomez2024multimodaldepressionvideo}. Some combine facial with acoustic or language features but omit rPPG~\cite{alghowinem2018multimodaldepression,zhou2026mddmarf,sadeghi2024llmfacialdepression}, while others include rPPG but omit eyegaze and head pose~\cite{kumar2025interpretabledepression,mu2026remotecognitivewellbeing}. We address these gaps with a multimodal deep learning model that predicts depression and anxiety in older adults with MCI from facial action units and landmarks, head pose, eyegaze, acoustic, language, and rPPG features collectively. 
 

\subsection{Cross-Modal Temporal Alignment}
Prior multimodal approaches handle temporal misalignment across features in one of two ways. The first approach temporally pools each feature time series into a single vector per feature before multimodal fusion~\cite{guo2022topicattentivedepression, jung2024hiquedepression, zhang2025interviewerbiasdepression}. The second approach keeps per-feature time series but places each feature observation at a single point on the time axis~\cite{fan2024transformermultimodaldepression, tao2024depmstat, gimenogomez2024multimodaldepressionvideo}, as illustrated in \autoref{fig:align_reason}(a). These methods do not provide a shared timeline that aligns frame-level, window-level, and word-level features according to their temporal support. In contrast, our temporal binning approach preserves the time interval covered by each feature by constructing a shared time axis that reflects the original recording times. This allows the model to learn temporally aligned cross-modal interactions, reflecting how clinicians assess patient behaviors in real-world clinical settings.

\subsection{Handling Missingness across Modality and Time}
Most prior multimodal mental health work handles missingness at the modality level for each sample, treating whether a modality is present for the participant as a whole~\cite{kumar2025interpretabledepression, chen2026scdmllm}. Other studies replace missing values at the timestep level with zeros, which cannot be distinguished from valid zero values in observed timesteps~\cite{kumar2025interpretabledepression, tao2024depmstat}. Gimeno-Gomez \textit{et al.}~\cite{gimenogomez2024multimodaldepressionvideo} encode missingness more finely, with per-modality, per-frame presence masks that act as attention masks for depression prediction. We build on this fine-grained masking strategy but place the masks on answer-aligned temporal bins. As a result, at each timestep, the multimodal features, their temporal availability masks, and the corresponding question context share the same timeline.

\subsection{Modeling Question and Answer Interaction}
\label{sec:related_work_qcond}
Prior work on mental health prediction in older adults with MCI ignores the question context in multimodal modeling and focuses only on the participant answer~\cite{mu2026remotecognitivewellbeing,zhou2022apathydepressionmci,zhou2023speechfacialmci}. In general-population depression classification, some methods condition the multimodal response on the question but in a limited way. Zhang \textit{et al.}~\cite{zhang2025interviewerbiasdepression} temporally pool the acoustic response time series and text into a single vector for each modality, then fuse them with the question using gated addition. This discards fine-grained temporal information. 
Other approaches apply question conditioning separately to each modality rather than to a joint multimodal representation. Niu \textit{et al.}~\cite{niu2021hcagdepression} use additive attention to condition the response acoustic time series and text sequence independently on the question before fusion. 
Guo \textit{et al.}~\cite{guo2022topicattentivedepression} condition only the text modality on the question context by encoding the concatenated question and response text with RoBERTa~\cite{liu2019roberta}, while the acoustic response is encoded independently before fusion. As a result, these methods do not condition the fused multimodal sequence on the question at a fine-grained temporal resolution within each answer. In addition, their conditioning incorporates only the acoustic and text modalities. In contrast, we add the question embedding to every timestep of the fused multimodal sequence, which also includes facial and physiological signals, allowing each moment of the answer to be interpreted in the context of the question that elicited it.

\begin{figure*}[t]
    \centering
    \includegraphics[width=\linewidth]{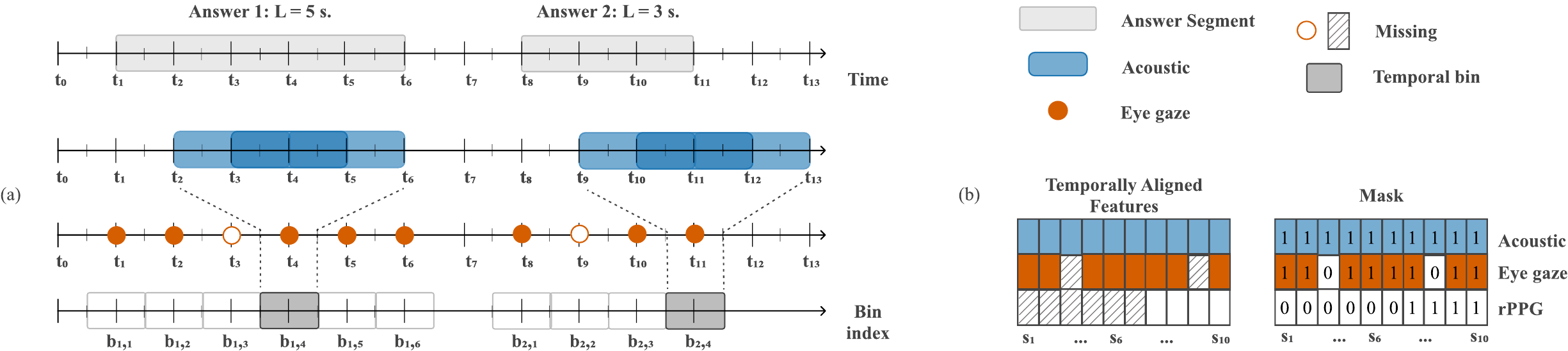}
    \caption{Adaptive temporal binning for within-answer multimodal alignment.
    (a) Each participant's recording is partitioned along the time axis into question-answer segments, as shown on the upper axis. Each segment is divided into a number of temporal bins depending on its duration, as shown on the lower axis. The bins define a shared timeline that aligns features of different temporal supports: frame-level features (e.g., eyegaze) are assigned to a bin by timestamp, and window-level features (e.g., acoustic) are assigned to every bin they overlap. (b) Bins from all answers are concatenated into a participant-level sequence, with a binary modality-timestep mask indicating observed and missing entries.}
    \label{fig:temporal_binning}
\end{figure*}

\subsection{Interpreting Predictions across Modality, Question, and Time}
Multimodal mental health interviews produce structured evidence across modalities, questions, and time, with multiple behavioral signals unfolding throughout the interview in response to specific prompts. Prior interpretability methods examine only a subset of these levels. Modality-level methods identify which modalities drive a prediction through performance comparisons~\cite{williamson2016multimodaldepression, jiang2024multimodaldigitalbiomarkers} or post-hoc attribution scores~\cite{kumar2025interpretabledepression, mu2026remotecognitivewellbeing}, without localizing importance to specific questions or moments in time. Question-level methods attribute importance to interview questions or to modalities within a question through attention~\cite{guo2022topicattentivedepression, jung2024hiquedepression} or performance comparisons~\cite{mandal2025questionwisefusion}, but do not identify when within an answer the relevant evidence is observed. Gimeno-Gomez \textit{et al.} apply Integrated Gradients (IG)~\cite{sundararajan2017integratedgradients} to visualize modality importance over time within a selected short temporal window from a single participant’s video rather than across the full interview~\cite{gimenogomez2024multimodaldepressionvideo}. They do not incorporate question structure or extend the analysis to cohort-level summaries for modality and temporal importance~\cite{gimenogomez2024multimodaldepressionvideo}. 

No prior method derives modality-level, question-level, and interview-wide temporal importance jointly across participants. We address this gap by deriving all three levels of explanation from IG~\cite{sundararajan2017integratedgradients} attributions at both cohort and individual levels over the full interview.

\section{Method}
\label{sec:method}

TAMI processes interviews in three stages: (1) temporally aligned multimodal feature time series with per-modality-feature per-timestep missingness masks; (2) question-conditioned multimodal feature fusion; and (3) post-hoc multi-level attribution that quantifies the importance of each modality, question, and timestep.

\subsection{Problem Formulation}
\label{sec:problem_formulation}

Each participant $i \in \{1,\ldots, N\}$ provides a remote interview recording represented as a sequence of observed question-answer segments, $\mathcal{X}_i = \{(q_{ij}, a_{ij})\}_{j=1}^{J_i},$ where $q_{ij}$ is the $j$-th question answered by participant $i$ and $a_{ij}$ is the corresponding answer. Because participants may not answer all interview questions, $J_i$ can vary across participants. Each answer spans a time interval of duration $L_{ij}$, which we partition into $B_{ij}$ temporal bins. The bins form a shared timeline that aligns features of different temporal supports. For modality-feature $m \in \mathcal{M}$ with feature dimension $D_m$, bin $b_{ij, k}$ with $k \in \{1, \ldots, B_{ij}\}$ corresponds to a feature vector $\mathbf{x}_{ij,k}^{(m)} \in \mathbb{R}^{D_m}$ and a binary availability mask $M_{ij,k}^{(m)} \in \{0,1\}$, where 0 indicates that the observation is missing. Given an interview, the task is to predict a binary label, representing the presence or absence of depression or anxiety. Separate models are trained for each task.

\subsection{Temporally Aligned Representation}

\subsubsection{Feature Extraction}
\label{sec:feature_extraction}
Throughout the paper, we use \emph{modality-feature} to refer to each of the eight feature time series used as model inputs. The features are extracted from the participant-side audiovisual recordings. From the facial video, we extract \textit{frame-level} features at 1 fps: combined facial action units and landmarks (AUs \& LMs), head pose~\cite{cheong2023pyfeat}, and eyegaze~\cite{abdelrahman2023l2csnet}. 
Physiological rPPG-derived heart rate is computed as a \textit{window-level} feature with a sliding window of 6s and 1s hop~\cite{boccignone2022pyvhr}. From the audio, we extract \textit{window-level} acoustic features with a sliding window of 2s and 1s hop: eGeMAPS, ComParE~\cite{eyben2010opensmile}, and Wav2Vec2 embeddings~\cite{baevski2020wav2vec2,wolf2019huggingface}. The language features are RoBERTa embeddings from WhisperX \textit{word-level} transcripts~\cite{bain2023whisperx,liu2019roberta}. Interview questions are encoded separately using Sentence-BERT~\cite{reimers2019sbert,wolf2019huggingface} and used to condition the participant-side features. These feature sets have been used in previous mental health~\cite{mu2026remotecognitivewellbeing, emden2026smartphonespeech, wagay2025redditmentalillness, jiang2024multimodaldigitalbiomarkers, jung2024hiquedepression} and behavioral analysis~\cite{dossantosmelicio2025adosbehavior, pabba2024visualclassroom} pipelines.
Full details are in the Supplementary in~\autoref{app:feature_extraction}.

\subsubsection{Answer Segmentation and Temporal Alignment}
\label{sec:alignment}
For this feasibility study, the interviewer manually marked the start and end timestamps of each question-answer interval. This step can potentially be automated at deployment by aligning the WhisperX transcript to the known interview questions to locate each question-answer interval. If a participant did not answer a question, that question is excluded for the participant. Each answer segment is divided into temporal bins with a bin width of $w=1$s, as illustrated in \autoref{fig:temporal_binning}(a). Because answer durations vary widely (1--926s), the number of bins adapts to each answer duration rather than truncating long answers or padding short ones. The number of bins for answer $j$ of participant $i$ is set as $B_{ij} = \mathrm{clamp}\left(\mathrm{round}(L_{ij}/w) + 1, B_{\min}, B_{\max}\right)$, where $B_{\min}$ and $B_{\max}$ are lower and upper bin-count bounds. These bounds ensure that every answered question is represented, while limiting the number of bins for long answers. When an answer exceeds the upper bound, the bin width is increased ($w>1$s) so that the full answer is represented using $B_{\max}$ bins without truncation. 


Frame-level features are assigned to the bin containing their frame timestamp. Window-level features are assigned to all bins that overlap their feature windows. If multiple windows overlap the same bin, their feature vectors are averaged to form a single bin-level representation, treating all overlapping windows as an estimate of the signal within that bin. Transcript words are assigned using WhisperX timestamps, and all words overlapping a bin are concatenated and encoded as a single bin-level language representation using RoBERTa embeddings. The bins from all answered questions are concatenated along the temporal axis as participant-level multimodal feature time series, shown in \autoref{fig:temporal_binning}(b), with a length of $S_i=\sum_{j=1}^{J_i} B_{ij}$, where bin index is $s\in\{1,\dots,S_i\}$ in the concatenated participant-level sequence.



\subsubsection{Modality-Timestep Missingness}
\label{sec:missingness}
The availability mask $M_{ij,k}^{(m)}$ for bin $b_{ij, k}$ is set to $0$ if no valid observation for feature $m$ is available in that bin. This occurs when no frame-level feature value falls within the bin interval, no window-level feature overlaps the bin interval, or no transcript word is spoken in the bin. In that case, the corresponding feature vector is zero-filled $\mathbf{x}_{ij,k}^{(m)}=\mathbf{0}$. Conversely, $M^{(m)}_{ij,k} = 1$ denotes a valid observation for that feature in the bin, even when the feature vector contains zero-valued measurements, thereby distinguishing missing observations from valid low-signal behavior (\autoref{fig:temporal_binning}(b)).

\subsection{Question-Conditioned Fusion and Prediction}
\label{sec:fusion_prediction}
\begin{figure}[t]
    \centering
    \includegraphics[width=\columnwidth,keepaspectratio]{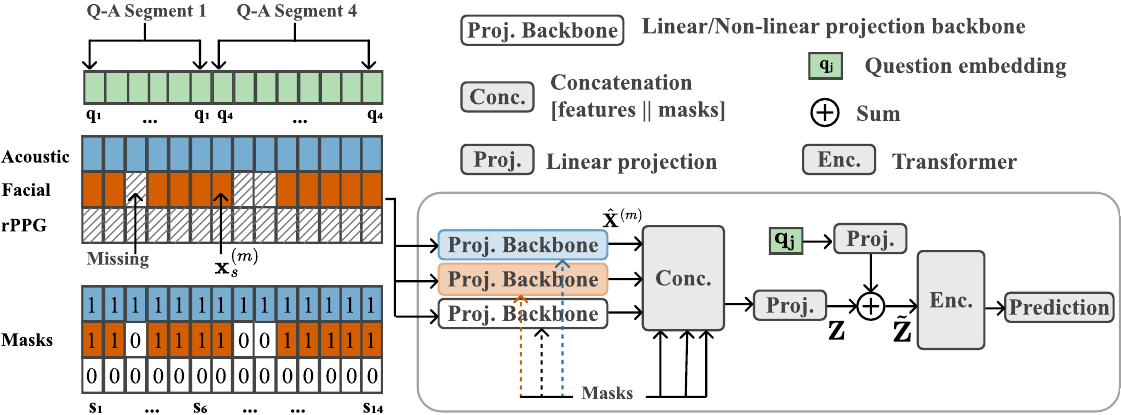}
    \caption{
Question-conditioned fusion pipeline ($S=14$ for illustration). 
} 
    \label{fig:fusion_backbones}
\end{figure}

\subsubsection{Projection Backbones and Fusion}
\label{sec:projection_backbones}
We explored a linear and a non-linear per-modality-feature projection backbone (\autoref{fig:fusion_backbones}) to test whether the model would benefit from learning a per-modality-feature temporal representation before multimodal fusion. The linear projection backbone projects each modality-feature at each temporal bin $s$ to dimension $d$ independently, with no temporal modeling (no mixing of information across time). Here, temporal modeling is deferred to the Transformer encoder (Enc.) shown in~\autoref{fig:fusion_backbones}, which applies attention across temporal bins after multimodal fusion. The non-linear projection backbone instead uses a per-modality-feature Transformer that applies attention across time before fusion. Thus, unlike the linear projection backbone, the non-linear projection backbone mixes information for each modality-feature across time before multimodal fusion.

Both backbones take the aligned per-modality-feature time series $\mathbf{X}_i^{(m)}\in\mathbb{R}^{S_i\times D_m}$ and availability masks $M_i^{(m)}\in\{0,1\}^{S_i\times 1}$ as input. Then they both project each modality-feature onto a common dimension $d$, producing $\hat{\mathbf{X}}_{i}^{(m)}\in\mathbb{R}^{S_i\times d}$: 

\begin{equation}
    \hat{\mathbf{X}}_{i}^{(m)} =
    \begin{cases}
        M_{i}^{(m)}\odot \big(\mathbf{X}_{i}^{(m)}\,W_m^{\top}\big)
            & \text{(a) linear}\\[4pt]
        f_m\!\big(\mathbf{X}_{i}^{(m)}\,W_m^{\top},\; M_{i}^{(m)}\big)
            & \text{(b) non-linear}
    \end{cases}
\end{equation}

The linear projection gates each feature by its mask. The non-linear projection instead applies a per-modality-feature Transformer $f_m$ with temporal self-attention while using the mask as the attention mask.
Regardless of which projection backbone is used, the resulting projected per-modality-feature representations $\hat{\mathbf{x}}_{i,s}^{(m)}\in\mathbb{R}^{d}$ at each bin $s$ are then concatenated with their availability masks~\cite{che2018grud}. Before concatenation, a zero-filled missing bin and an observed value at the feature mean are both zero, so without appending the mask as a missingness indicator, the model cannot distinguish a missing observation from a valid zero measurement. The concatenated representation and mask vectors are then projected to a single fused multimodal token $\mathbf{z}_{i,s}\in\mathbb{R}^{d}$:
$    \mathbf{z}_{i,s}
    = W_f \big[\,
        \hat{\mathbf{x}}_{i,s}^{(1)};\cdots;\hat{\mathbf{x}}_{i,s}^{(N_m)};\;
        M_{i,s}^{(1)};\cdots;M_{i,s}^{(N_m)}
      \,\big],$ where $N_m = 8$ is the total number of modality-features. 
The tokens are stacked to form the fused multimodal sequence $\mathbf{Z}_{i}\in\mathbb{R}^{S_i\times d}$. 

\subsubsection{Question Conditioning and Prediction}
\label{sec:q_cond_pred}
We condition the fused multimodal sequence, $\mathbf{Z}_{i}$, on the question context at each temporal bin. Let $\mathbf{q}_{ij}$ be the embedding of the $j$-th question for participant $i$. For each temporal bin $s$ within the answer to question $j$, we add the question embedding to the fused multimodal token $\mathbf{z}_{i,s}$,
    $\tilde{\mathbf{z}}_{i,s}
    = \mathbf{z}_{i,s} + W_q\, \mathbf{q}_{ij}$, where $W_q$ maps the question embedding to dimension $d$. The conditioned tokens form a sequence
$\tilde{\mathbf{Z}}_{i}\in\mathbb{R}^{S_i\times d}$, which is then processed by a Transformer encoder (Enc. in~\autoref{fig:fusion_backbones}) with a prepended \texttt{[CLS]} token, and a linear head maps the \texttt{[CLS]} representation to the prediction logit. 
Within a batch, participant sequences of unequal length are padded to the longest and excluded from attention through a padding mask.



%
\begin{figure}[t]
    \includegraphics[width=\columnwidth,keepaspectratio]{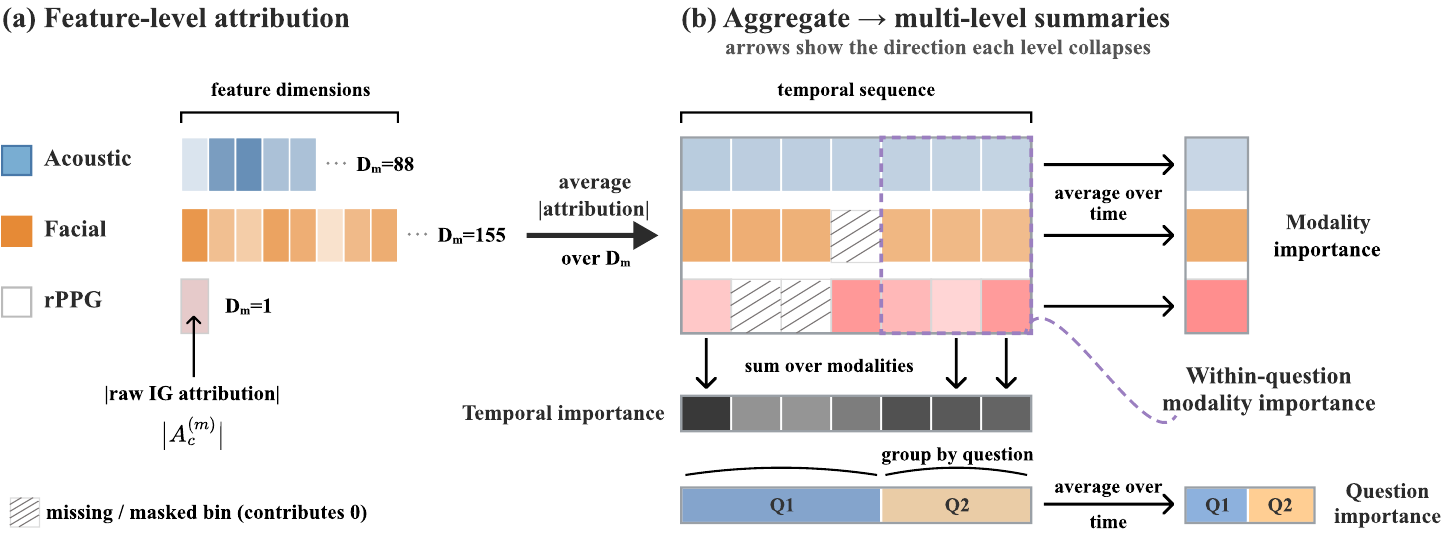}
    \caption{Multi-level interpretability in modality, question, and time. 
}
    \label{fig:attr}
\end{figure}

\subsection{Post-hoc Multi-Level Attribution}
\label{sec:interpretability}

After training, we attribute each prediction to the input features using IG, with a zero-vector baseline equal to the mean feature value after standardization~\cite{sundararajan2017integratedgradients}. This yields a single scalar attribution $|A_{c,s}^{(m)}|$ for each feature $c$ of modality-feature $m$ at each temporal bin $s$ for each participant (\autoref{fig:attr}(a)). We aggregate these feature-level attributions to build interpretable summaries along clinically interpretable axes: which question and which modality within that question drove the prediction, and when in the interview the relevant evidence occurred (\autoref{fig:attr}(b)). 
The arrows in \autoref{fig:attr}(b) show these aggregations, producing modality importance, question importance, within-question modality importance, and the temporal importance across the interview. Full derivations are provided in the Supplementary in~\autoref{app:interpretability}.

\section{Experiments}

\subsection{Dataset}
\label{sec:data}
Our dataset comprises remote semi-structured interviews from 49 older adults with MCI. Interviews were conducted over Zoom at participants’ preferred time and location, typically at home, using their own devices, including tablets, laptops, and mobile phones. \ifanonymous
The study protocol was approved by the relevant institutional review board.
\else
Participants were recruited from the Charlie and Harriet Shaffer Cognitive Empowerment Program (CEP) at Emory University. The study protocol was approved by Emory University Institutional Review Board
(protocol \#2025P012652).
\fi
Participants provided written informed consent for research participation. Each interview included six question groups: open-ended narrative prompts including a TAT picture description task \cite{morgan1935tat}, an I-CONECT social and functional check \cite{yu2021iconect}, the Geriatric Depression Scale (GDS) \cite{yesavage1982gds}, the Generalized Anxiety Disorder (GAD-7) scale \cite{spitzer2006gad7}, the UCLA Loneliness Scale \cite{russell1978uclaloneliness}, and the Lubben Social Network Scale (Social Engagement) \cite{lubben1988lsns}. Before each interview, participants were told that they could elaborate on any answer. However, answers to yes/no and Likert-scale questions were usually brief. Not every participant answered every question, owing to question skip logic, pilot-phase changes to the administered question groups, and occasional technical issues (details are provided in the Supplementary in~\autoref{app:dataset}).

Participants had a mean age of 73.4 $\pm$ 8.1 years, and 47\% were female. Across all questions, answers had a median duration of 5s with a maximum duration of 926s $\approx 15$ minutes (interquartile range: 3--12s). Full interview durations (including both participant and interviewer sides) ranged from 21 to 67 minutes, with a median of 31 minutes. Participant-side video resolution varied from $640\times480$ to $1920\times1080$ pixels. We define depression risk as GDS $\ge$ 10, corresponding to at least mild depressive symptoms~\cite{yesavage1982gds} and anxiety risk as GAD-7 $\ge$ 5, corresponding to at least mild anxiety symptoms~\cite{spitzer2006gad7}. One participant lacked a GAD-7 score, yielding 49 participants for depression (21 positive, 28 negative) and 48 for anxiety (17 positive, 31 negative). Per-modality-feature average temporal bin missingness was at most 3\% except for language (37\%). Structured question items usually elicit single-word or yes/no answers (median duration: 3s), so most one-second bins within an answer contain no transcribed word, leading to high language-feature missingness (average $\approx 44\%$ across structured question groups), while open-ended questions elicit the longest answers (median duration: 87.5s) and the lowest missingness (16\%). The Supplementary reports language-feature missingness and answer duration by question group in~\autoref{fig:txt}.




\subsection{Implementation and Evaluation} 
\label{sec:implementation_eval}

Temporal bins are constructed with a minimum of $B_{\min}=1$
bin and a maximum of $B_{\max} = 64$ bins per answer. The upper bound of 64 bins was chosen so that the majority of answers are represented at 1-second resolution. For the 4\% of answers exceeding 64 seconds, bins were widened proportionally so that the full answer duration was still spread across all 64 bins, reducing temporal resolution rather than discarding content (\autoref{sec:alignment}). All Transformer encoders use a hidden dimension $d=128$. The per-modality-feature encoders in the non-linear projection backbone use a single Transformer layer, while the common encoder (the Enc. in~\autoref{fig:fusion_backbones}) in both backbones uses two Transformer layers. We train all models for 10 epochs (sufficient for plateauing the model training loss curve)~\cite{mu2026remotecognitivewellbeing} using the Adam optimizer with a learning rate of 0.001, a batch size of 4 participants, and binary cross-entropy loss. A class-balanced batch sampler is used to address class imbalance. Interpretability analysis is applied to the best-performing model variant for each task (Non-linear+T for depression and Linear+T for anxiety). 


We evaluated all models using participant-independent 5-fold cross-validation across 3 repetitions (15 runs). All feature time series were standardized using training-fold statistics before training to have zero mean and unit variance. We report AUROC scores with 95\% confidence intervals (CIs) across all folds. We also report area under the precision-recall curve (AUPRC), macro-F1, precision, and recall scores for all models in the Supplementary in~\autoref{tab:extended_classification_results}. All performance comparisons use one-sided paired $t$-tests on fold-level AUROC score differences at $p<0.05$.


\subsection{Ablations} 
\label{sec:ablation}
We evaluate temporal alignment (T) and question-conditioned fusion (Q) of the multimodal feature time series as two design options under both linear and non-linear projection backbones. The sequence-index-paired baseline model pairs feature time series by index rather than by the timestamp of the underlying behavior, following prior multimodal models~\cite{fan2024transformermultimodaldepression, tao2024depmstat}. Models that use Q only without T test two adaptations of prior question-conditioning methods on the temporally unaligned sequence-index-paired baseline. The first is \textit{pooled additive conditioning} ($+$ Q, sum), adapted from the question conditioning approach of Zhang \textit{et al.}~\cite{zhang2025interviewerbiasdepression}. In our variant, all answered-question embeddings are averaged into one interview-level vector and added uniformly across time to the fused multimodal sequence. The second is \textit{cross-attention conditioning} ($+$ Q, cross-attn.), which generalizes the question-conditioned additive attention used in HCAG~\cite{niu2021hcagdepression}. In our variant, cross-attention is instead applied between the fused multimodal sequence and the sequence of answered-question embeddings. Further details are provided in the Supplementary in~\autoref{app:ablations}. 

We also ablate the modality-timestep missingness mask on the T models with both linear and non-linear projection backbones. We replace the mask with an all-ones mask (no mask variant), which makes zero-filled bins indistinguishable from observed ones, reproducing the no-mask case used by prior work that zero-fills at the timestep level~\cite{kumar2025interpretabledepression, tao2024depmstat}.



\section{Results}
\label{sec:results}

\begin{table}[t]

\centering

\caption{
AUROC performance (mean $\pm$ 95\% CIs) for depression and anxiety classification.
\textbf{Bold}, \underline{underline}, and \textnormal{\textit{italics}} denote the best, second-best, and third-best results for each task, respectively. Tied results receive the same rank.
T denotes temporally aligned features, and Q denotes question conditioning. Linear and Non-linear denote the different projection backbones.
}

\label{tab:main_results}

\footnotesize
\setlength{\tabcolsep}{4pt}
\renewcommand{\arraystretch}{1.08}

\begin{tabular*}{\columnwidth}{@{\extracolsep{\fill}}lcc@{}}
\toprule

\textbf{Variant}
& \makecell[c]{\textbf{Depression}\\\textbf{AUROC} $\uparrow$}
& \makecell[c]{\textbf{Anxiety}\\\textbf{AUROC} $\uparrow$} \\

\midrule

\multicolumn{3}{@{}l}{\textit{Sequence-index-paired baseline} \cite{fan2024transformermultimodaldepression,tao2024depmstat}} \\

Linear
& $0.51 \pm 0.06$
& $0.58 \pm 0.03$ \\

Non-linear
& $0.58 \pm 0.11$
& $0.56 \pm 0.06$ \\

\midrule


Linear + T
& $\mathit{0.64 \pm 0.08}^{\ast}$
& $\mathbf{0.69 \pm 0.09}^{\P}$ \\

Non-linear + T
& $\mathbf{0.68 \pm 0.04}$
& $0.58 \pm 0.07$ \\

\midrule


Linear + Q, sum~\cite{zhang2025interviewerbiasdepression}
& $0.52 \pm 0.003$
& $0.55 \pm 0.06$ \\

Linear + Q, cross-attn.~\cite{niu2021hcagdepression}
& $0.45 \pm 0.07$
& $0.56 \pm 0.04$ \\

Non-linear + Q, sum~\cite{zhang2025interviewerbiasdepression}
& $0.55 \pm 0.08$
& $0.54 \pm 0.07$ \\

Non-linear + Q, cross-attn.~\cite{niu2021hcagdepression}
& $0.56 \pm 0.11$
& $0.56 \pm 0.09$ \\

\midrule


Linear + Q + T
& $\underline{0.67 \pm 0.10}^{\dagger}$
& $\underline{0.67 \pm 0.07}^{\S}$ \\

Non-linear + Q + T
& $\underline{0.67 \pm 0.07}^{\ddagger}$
& $\mathit{0.61 \pm 0.01}$ \\

\bottomrule
\end{tabular*}

{\footnotesize
\raggedright

Selected AUROC comparisons using one-sided Student's paired $t$-tests:

Depression: $^{\ast}$Linear+T $>$ Linear baseline, $p\approx0.05$;
$^{\dagger}$Linear+Q+T $>$ Linear+Q (sum), $p\approx0.009$ and $>$ Linear+Q (cross-attn.), $p\approx0.001$, and $>$ Linear baseline, $p\approx0.009$;
$^{\ddagger}$Non-linear+Q+T $>$ Non-linear+Q (sum), $p\approx0.02$, and $>$ Non-linear+Q (cross-attn.), $p\approx0.001$.

Anxiety: $^{\P}$Linear+T $>$ Linear baseline, $p\approx0.05$;
$^{\S}$Linear+Q+T $>$ Linear+Q (sum), $p\approx0.04$. For all other pairwise comparisons, $p>0.05$. 
Full pairwise statistical comparisons are provided in the Supplementary in~\autoref{tab:appendix_pairwise_pvalues}.
Macro-F1, AUPRC, precision, and recall results are provided in the Supplementary in~\autoref{tab:extended_classification_results}.

\par
}

\end{table}

\begin{table}[t]

\centering

\caption{
Mask ablation results for depression and anxiety classification in terms of AUROC (mean $\pm$ 95\% CIs). \textbf{Bold} denotes the better result within each projection backbone and task.
}

\label{tab:mask_ablation}

\footnotesize
\setlength{\tabcolsep}{4pt}
\renewcommand{\arraystretch}{1.08}

\begin{tabular*}{\columnwidth}{@{\extracolsep{\fill}}lcc@{}}
\toprule

\textbf{Variant}
& \makecell[c]{\textbf{Depression}\\\textbf{AUROC} $\uparrow$}
& \makecell[c]{\textbf{Anxiety}\\\textbf{AUROC} $\uparrow$} \\

\midrule

Linear + T (no mask)
& $\mathbf{0.65 \pm 0.09}$
& $0.64 \pm 0.05$ \\

Linear + T (mask)
& $0.64 \pm 0.08$
& $\mathbf{0.69 \pm 0.09}$ \\

\midrule

Non-linear + T (no mask)
& $0.67 \pm 0.09$
& $0.56 \pm 0.09$ \\

Non-linear + T (mask)
& $\mathbf{0.68 \pm 0.04}$
& $\mathbf{0.58 \pm 0.07}$ \\

\bottomrule
\end{tabular*}

\end{table}

\subsection{Overall Performance}
\label{sec:main_res}
\autoref{tab:main_results} reports AUROC scores for depression and anxiety under each combination of projection backbone (Linear vs. Non-linear), temporal alignment (T), and question conditioning (Q). Temporally aligning the multimodal features achieves the highest scores for both tasks: Non-linear+T reaches $0.68 \pm 0.04$ for depression ($\Delta{=}{+}0.10$ over the Non-linear baseline), and Linear+T reaches $0.69 \pm 0.09$ for anxiety ($\Delta{=}{+}0.11$ over the Linear baseline, $p{=}0.05$). Adding question conditioning on top of alignment provides no significant gain over alignment alone (all Q+T vs.\ T $p>0.05$), though Q+T variants remain competitive (Linear +Q+T: $0.67$ for depression and anxiety). 
Without temporal alignment, neither summation nor cross-attention for question conditioning shows a consistent advantage across the two tasks (AUROC 0.45--0.56). Models with linear and non-linear projection backbones have overlapping confidence intervals, and their rankings are inconsistent across tasks. Replacing the modality-feature-timestep mask with all-ones, thereby treating all bins as observed (the no-mask variant in ~\autoref{tab:mask_ablation}), reduces performance for anxiety classification (Linear $\Delta=-0.05$ and Non-linear $\Delta=-0.02$). The effect on depression classification is smaller, where removing the mask increases AUROC by 0.01 for the linear model but decreases it by 0.01 for the non-linear model.

\subsection{Modality Importance}
\label{sec:modality_importance}

\begin{figure*}[t]
    \centering
    \includegraphics[width=\linewidth]{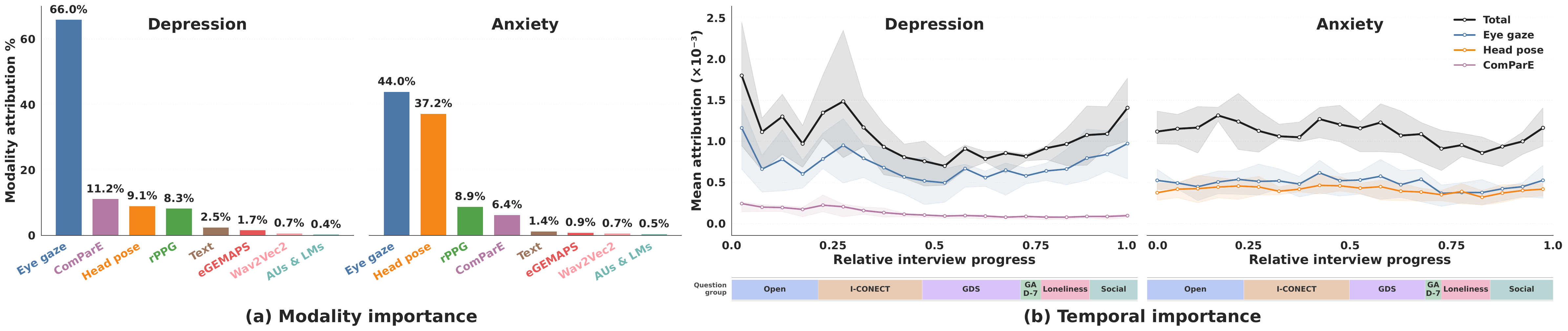}
    \caption{IG attribution for depression and anxiety across participants. (a) Modality importance across the entire video duration. 
    (b) Temporal importance across interview progress, showing the total attribution together with the eyegaze, head pose, and ComParE time series attributions. Shaded bands indicate participant-level variability, and the colored strip below each plot marks the interview progress of each question group. 
    }
    \label{fig:ig_main}
\end{figure*}

The most important features differ across the two tasks (\autoref{fig:ig_main}(a)). For depression, eyegaze accounts for 66\% of the total attribution, followed by ComParE at 11.2\%. For anxiety, the attribution is more evenly distributed between eyegaze at 44\% and head pose at 37.2\%. Together, the two leading features account for more than 77\% of the total attribution in each task. AUs \& LMs contribute less than 1\% in both tasks, indicating that eyegaze and head pose capture most of the predictive facial signal once they are present. Language features contribute less than 3\% in both tasks. Acoustic information is more prominent for depression, with ComParE ranking second overall, whereas no individual acoustic feature exceeds 6.5\% for anxiety. This feature dominance holds at the question level for both tasks, as shown for the highest-ranked questions in the Supplementary in~\autoref{fig:ig_question_m}.

\begin{table}[t]
\centering
\caption{Question-group IG attribution summaries. Values are scaled by $10^3$. Spread is the difference between the maximum and minimum group means within each task.
\textbf{Bold}, \underline{underline}, and \textnormal{\textit{italics}} denote the highest, second-highest, and third-highest
attributions across question groups within each task.
}
\label{tab:question_group_importance}
\footnotesize
\setlength{\tabcolsep}{3pt}
\renewcommand{\arraystretch}{1.08}
\begin{tabular*}{\columnwidth}{@{\extracolsep{\fill}}l cc cc@{}}
\toprule
& \multicolumn{2}{c}{\textbf{Depression}} & \multicolumn{2}{c}{\textbf{Anxiety}} \\
\cmidrule(lr){2-3} \cmidrule(lr){4-5}
\textbf{Question group} & Range & Mean & Range & Mean \\
\midrule
Open              & 1.09--1.65 & \textbf{1.31}    & 1.04--1.39 & \underline{1.16} \\
I-CONECT          & 0.66--2.15 & \textit{1.03}    & 0.91--1.58 & \textbf{1.17} \\
GDS               & 0.42--1.03 & 0.68             & 0.74--1.76 & \textbf{1.17} \\
GAD-7             & 0.73--1.07 & 0.90             & 0.80--1.17 & 1.03 \\
Loneliness        & 0.64--1.35 & 0.88             & 0.77--1.14 & 0.93 \\
Social engagement & 0.83--1.70 & \underline{1.17} & 0.74--1.64 & \textit{1.10} \\
\midrule
\textit{Spread}   & \multicolumn{2}{c}{0.63} & \multicolumn{2}{c}{0.24} \\
\bottomrule
\end{tabular*}
\end{table}

\subsection{Question Importance}
\label{sec:question_importance}

Both tasks distribute attribution across all interview questions but differ in how that attribution spreads across question groups (\autoref{tab:question_group_importance}). Depression shows wider dispersion (spread: $0.63 \times 10^{-3}$), with the highest mean attribution assigned to the open-ended questions, followed by the social engagement questions. The open-ended questions are Q1 (``Tell me about yourself''), Q2 (Best event in the past two years), Q3 (Worst event in the past two years), and Q4 (TAT picture description). At the individual question level, the four questions with the highest absolute attributions are Q10 (I-CONECT ``In the past week, have you had visitors who stayed with you in your home for a night or more? If yes, then how many nights?''), Q89 (Social Engagement: ``How often is one of your relatives available for you to talk to when you have an important decision to make?''), Q5 (I-CONECT: ``So, would you say your health is Very good, Good, Fair, Poor, Excellent?''), and Q1 (open-ended: ``Tell me about yourself'').

Anxiety shows a tight group-mean range (spread: $0.24 \times 10^{-3}$), indicating roughly uniform reliance on the full interview, with the highest-attribution questions coming from multiple groups. The four questions with the highest attribution for anxiety are Q47 (GDS: crying frequency), Q91 (Social Engagement: ``How often is one of your friends available for you to talk to when you have an important decision to make?''), Q90 (Social Engagement: ``When one of your friends has an important decision to make, how often do they talk to you about it?''), and Q20 (I-CONECT: ``Did you spend time communicating with any friends or family members in writing, such as email, text, or letter writing this week?''). The attribution scores for the highest-ranked questions across all question groups are presented for each task in the Supplementary in~\autoref{fig:ig_question_q}.

Neither GDS nor GAD-7 ranks among the top question groups for its corresponding task, indicating that the model relies on participant behavioral responses across the video rather than on the screening tool itself. This suggests a novel opportunity for developing multimodal behavioral biomarkers associated with depression and anxiety in MCI beyond a brief self-report tool for clinical decision support. 


\subsection{Temporal Importance}
\label{sec:temporal_importance}

Depression attribution peaks early and decays across the interview from $\sim$2.0 to $\sim$0.75 ($\times 10^{-3}$) by mid-interview (\autoref{fig:ig_main}(b)). The decline is driven by eyegaze, whose contribution decreases from $\sim$1.25 to $\sim$0.50 ($\times 10^{-3}$) over the same period. The early peak aligns with the open-ended questions and is consistent with the question-level finding that these questions receive the highest mean attribution. In contrast, anxiety shows a uniform temporal importance profile. Total attribution decreases from $\sim$1.2 to $\sim$0.90 ($\times 10^{-3}$) near the end of the interview before returning to $\sim$1.2. The eye gaze and head pose attributions remain relatively uniform throughout the interview. Representative participant profiles are shown in the Supplementary in~\autoref{fig:single_patient_profiles}.

\subsection{Open-Ended Question Performance}
\label{sec:open_perf}
To evaluate whether the open-ended portion of the interview alone retains predictive performance, we train and evaluate the best-performing model for each task using only multimodal features extracted from participants’ responses to the four open-ended questions (median duration: 5.1 min). We use the Non-linear+T model for depression and the Linear+T model for anxiety. 
For depression, the open-ended-only model achieves an AUROC of $0.67 \pm 0.13$, which is only 0.01 lower than that of the model trained on the full interview. This difference is not statistically significant $(p>0.05)$. This result is consistent with the question importance analysis, which shows that depression predictions rely on responses to the open-ended questions. For anxiety, the open-ended-only model achieves an AUROC of $0.59 \pm 0.06$. The model trained on the full interview achieves an AUROC that is 0.10 higher, and the difference is statistically significant ($p\approx0.02$). This finding is also consistent with the question importance analysis, which suggests that anxiety predictions rely uniformly on the full interview.

\section{Discussion}

\subsection{Temporal Alignment}
Temporally aligning the multimodal features was the primary driver of performance across both tasks, with models that used temporal alignment achieving the highest AUROC scores compared to the baseline models. Aligning features by timestamp corrects the temporal misalignment between the multimodal behavioral time series caused by sequence-index pairing used in prior approaches~\cite{tao2024depmstat, fan2024transformermultimodaldepression} (\autoref{fig:align_reason} and \autoref{fig:alignment_missingness_examples}(a)).

\subsection{Question Conditioning}
Question-conditioned fusion did not improve the performance of models that used temporally aligned multimodal features, and the temporally unaligned Q-only model variants showed no consistent gains over the baselines on either task. Question conditioning injects semantic context about what was asked, which is directly relevant to interpreting the answer content. However, because language alone contributed less than 3\% of total attribution in both tasks, question content may have had limited influence on the prediction. Moreover, the Q-only variants do not use the explicit question-answer mapping provided by temporal alignment, so the model must recover this mapping, which may be difficult to learn in a data-driven manner.

\subsection{Missingness Awareness}

Overall, the modality–feature-timestep mask improved AUROC in three of the four model-task combinations, with the largest numerical gain for the Linear+T anxiety model and small, mixed effects for the depression models (\autoref{tab:mask_ablation}). To explore whether label-associated missingness might contribute to this pattern, we followed Che \textit{et al.}~\cite{che2018grud} and computed Pearson correlations $(r)$ across participants between modality missingness rates and the binary label for each task (depression or anxiety). We considered missingness informative when its rate correlated with the label and could therefore carry predictive information~\cite{che2018grud}. Based on the modality importance analysis, we examined this missingness-label association for the highest-ranked modalities in each task: eyegaze and head pose for anxiety prediction, and eyegaze for depression prediction. We also included text as a higher-missingness reference for anxiety prediction. Missingness in eyegaze and head pose had absolute correlations of $|r|=0.19$ and $0.18$, respectively, with the anxiety label, compared with $|r|=0.04$ for text (highest-missing modality). This suggests that missingness in the dominant anxiety modalities was informative and that the much higher rate of text missingness was not as informative. For depression, the correlation between eyegaze missingness and the label was small ($|r|=0.014$), suggesting that missingness was not informative for this task.\footnote{These magnitudes are comparable to those reported in Fig. 1 in~\cite{che2018grud}, where the missingness–label correlations were below 0.3.} This task-specific pattern is consistent with the finding reported by Che \textit{et al.}~\cite{che2018grud} that modeling missingness can improve prediction when it is associated with the label. In contrast, missingness with little or no label association provides limited benefit. This may explain why the mask numerically improved both anxiety models but had small, mixed effects on the depression models. However, the performance differences were not statistically significant at this sample size ($p>0.05$), and this interpretation requires validation in a larger sample. The mask is also important for the interpretability analysis because, without it, the modality, question, and temporal importance summaries could assign attribution to behavior that was not observed.

\subsection{Modality Importance}
Eyegaze was the dominant feature for both depression and anxiety, although anxiety distributed attribution more evenly between eyegaze and head pose. For depression, eyegaze accounted for 66\% of the total attribution, consistent with findings of eye movement differences between individuals with and without depression, including longer saccade durations~\cite{takahashi2021eyemovementmdd}. Among older adults specifically, depression was associated with fewer fixations and saccades~\cite{takahashi2021eyemovementmdd}. The ComParE acoustic feature set ranked second for depression with 11.2\% attribution, supporting previous evidence that depression is associated with distinct vocal characteristics such as monotony, reduced pitch variability, and slower speech~\cite{cummins2015speechdepressionsuicide}. For anxiety, eyegaze and head pose accounted for 44\% and 37.2\% of the total attribution, respectively. This is consistent with reduced face-directed gaze and shorter fixations in socially anxious individuals during real-time interactions~\cite{kim2018socialanxietygaze, konovalova2021socialanxietygaze}, as well as increased head movement and velocity during anxiety and stress states in video-based facial analysis~\cite{giannakakis2017facialstressanxiety}. All remaining acoustic and language features accounted for less than 6.5\% of the attribution in both tasks, consistent with previous work on quantifying psychological well-being in older adults with MCI~\cite{mu2026remotecognitivewellbeing}.


\subsection{Question Importance}
The most notable result from the interpretability analysis is that the model did not assign its highest attribution to the question group that defined each label. GDS ranked last of six question groups for depression, and GAD-7 ranked second to last for anxiety. Attribution was instead distributed across question groups, suggesting that the model relied on participant behavior throughout the interview rather than on the screening questions. At the individual-question level, the highest-attributed questions for depression concerned social contact and support, and self-rated health. This is consistent with evidence that smaller social networks and lower social participation are associated with depressive symptoms in older adults~\cite{wendel2022socialnetworkdepression}, as well as evidence linking self-rated health to depressive symptoms in this population~\cite{peleg2021selfratedhealthdepression}. For anxiety, crying frequency received the highest attribution, which is consistent with evidence that greater attachment anxiety is associated with longer and more intense crying episodes~\cite{millings2016cryingproneness}. Tearfulness is also included under the Tension item of the Hamilton Anxiety Rating Scale~\cite{maier1988hamiltonanxiety}. The next highest-attributed questions concerned communication with and support from friends, which is consistent with evidence that low social support and social isolation are among the most reported factors associated with anxiety symptoms in older adults~\cite{shafiee2025anxietyolderadults}.

\subsection{Temporal Importance}
Depression and anxiety differed in where the model’s attributions were concentrated. For depression, attribution was highest during the open-ended question block and declined across the later structured section. This is consistent with prior work that reported higher depression prediction performance from spontaneous speech than from a constrained sentence-reading task~\cite{alghowinem2013spontaneousread}. This suggests that spontaneous-speech response formats are more informative for depression screening. Anxiety, instead, showed a uniform spread of attribution across question groups. The most informative features, eyegaze and head pose, were expressed continuously rather than elicited by a particular question. This is consistent with accounts of anxious hypervigilance as sustained monitoring of the surroundings through eye movements that occurs independently of the immediate stimulus~\cite{richards2014hypervigilance}, indicating that anxiety-related signals reflect a continual behavioral state.

\subsection{Open-Ended Question Performance}

Restricting the model input to participants’ responses to the four open-ended questions yielded a depression classification AUROC of 0.67, only 0.01 below the 0.68 achieved using the full interview, with no statistically significant difference between them ($p>0.05$). Although combined open-ended responses had a median duration of 5.1 minutes (range: 2.6--11.5 min) for participants who answered all the open-ended questions ($n=25$), the short responses retained predictive information. This is consistent with the high attribution assigned to the open-ended question group from the interpretability analysis. For anxiety, however, restricting the input to the open-ended questions significantly reduced AUROC from 0.69 to 0.59 ($p\approx0.02$). This result is consistent with the interpretability analysis, which indicates that anxiety-related information is distributed across the full interview. None of the four questions with the highest anxiety attribution belonged to the open-ended question group, suggesting that an abbreviated anxiety protocol may require questions selected from other parts of the interview. Given this cohort's median participant-side-only interview duration of 19 minutes (maximum 60 min), a shorter protocol centered on open-ended questions (median duration: 5.1 min; maximum: 11.5 min) could reduce clinician time and participant burden for depression assessment. Our findings suggest that using the four open-ended questions alone may be sufficient to screen for depression. However, this finding requires validation in a larger sample.



\subsection{Limitations and Future Work} 
One major limitation of this study is the small sample size ($N=49$).
In addition, the feature extractors for facial and acoustic features were pretrained on general-population data and may be less reliable on remote recordings from older adults with MCI, as reported in previous studies~\cite{mu2026remotecognitivewellbeing}. In future work, our findings should be validated in larger and more diverse cohorts, including populations with neuropsychiatric conditions who often have symptoms associated with depression and anxiety.

\section{Conclusion}
Depression and anxiety are often underdiagnosed in older adults with MCI. Current ML approaches can analyze multimodal behavioral signals for screening. However, these approaches overlook the importance of cross-modal temporal alignment, fail to account for time-varying modality missingness and question context, and provide limited interpretability for clinicians. We presented TAMI, a multimodal framework for detecting depression and anxiety in older adults with MCI from remote interviews, which incorporates cross-modal temporal alignment, modality-timestep missingness handling, question conditioning, and multi-level interpretability. The framework achieved 0.68 (depression) and 0.69 (anxiety) AUROC scores, with temporal alignment providing the largest performance gain over baselines.   
When restricted to multimodal responses from the four open-ended questions, the model achieved an AUROC score of 0.67 for depression. These responses had a median duration of 5.1 minutes, compared with 19 minutes for the participant-side full interview.
This suggests that a shorter open-ended protocol may be sufficient to screen for depression. The proposed temporal alignment approach may benefit other applications involving multimodal time series, including emotion recognition~\cite{wu2025multimodalemotionrecognition} and human activity recognition~\cite{karim2025humanactionrecognition}, where features differ in sampling rates and temporal support, and cross-modal temporal correspondence is important for prediction. Clinically, TAMI supports remote mental health screening for adults who face barriers to in-person care or live in underserved regions. Its multi-level interpretability (modality, question, and temporal attributions) also allows clinicians to understand the reasoning underlying the model's predictions, supporting informed clinical review.


\bibliographystyle{IEEEtran}
\bibliography{TAMI_references}







\clearpage

\renewcommand{\appendixname}{Supplementary}

\appendices


\section{Method}
\label{app:method}

\subsection{Feature Extraction}
\label{app:feature_extraction}

We extract four modalities from the participant-side video recording: facial, acoustic, language, and physiological modalities. The eight modality-feature time series used as model inputs are action units and landmarks, head pose, eyegaze, eGeMAPS, ComParE, Wav2Vec2, rPPG-derived heart rate, and transcript embeddings.~\autoref{tab:features} summarizes these inputs.

For the extracted features, we define window-level features as features that summarize signal content over an interval (e.g., acoustic or physiological features extracted with sliding windows), and frame-level features as features that are measured at individual timestamps (e.g., facial features extracted at 1 frame per second (fps)).

\subsubsection{Facial video features}
Frame-level features are extracted at 1~frame per second (fps) from recordings originally captured at 25 fps and include facial action units and landmarks ($D_a = 155$), head pose ($D_h = 3$), and eyegaze
($D_e = 2$), using the Py-Feat library \cite{cheong2023pyfeat} and L2CS-Net \cite{abdelrahman2023l2csnet}. 
For frames in which face detection or gaze estimation fails because of camera viewpoint, internet connectivity, or lighting, the corresponding temporal bins (defined in~\autoref{sec:alignment}) are marked missing.

\begin{table}[!b]
\centering
\caption{Model-input features.}
\label{tab:features}
\footnotesize
\setlength{\tabcolsep}{4pt}
\renewcommand{\arraystretch}{1.05}
\begin{tabular}{@{}lllr@{}}
\toprule
\textbf{Feature} & \textbf{Modality} & \textbf{Type} & \textbf{$D_m$} \\
\midrule
AUs \& landmarks & Facial & Frame-level & 155 \\
Head pose & Facial & Frame-level & 3 \\
Eyegaze & Facial & Frame-level & 2 \\
eGeMAPS & Acoustic & Window-level & 88 \\
ComParE & Acoustic & Window-level & 6{,}373 \\
Wav2Vec2 & Acoustic & Window-level & 1{,}024 \\
RoBERTa transcript & Language & Bin-level & 768 \\
rPPG heart rate & Physiological & Window-level & 1 \\
\bottomrule
\end{tabular}
\end{table}

\subsubsection{Acoustic features}
Window-level features are extracted using 2-second windows with 1-second overlap. Features include eGeMAPS ($D_g = 88$) and ComParE ($D_c = 6{,}373$) extracted using openSMILE~\cite{eyben2010opensmile},
and Wav2Vec2 embeddings ($D_w = 1{,}024$)~\cite{baevski2020wav2vec2} extracted using 
the Hugging Face Transformers library~\cite{wolf2019huggingface} from participant-side audio.

\subsubsection{Language features}
Participant-side audio is segmented into 30-second windows with 5-second overlap, then transcribed using WhisperX~\cite{bain2023whisperx} to obtain
word-level timestamps. Words whose timestamps fall within a temporal bin (defined in~\autoref{sec:alignment}) are concatenated and encoded with \texttt{roberta-base} \cite{liu2019roberta, jung2024hiquedepression}, yielding a $D_t = 768$-dimensional feature vector per bin, which we define as a bin-level feature.

Interview questions are encoded with a sentence-level encoder (Sentence Transformer \cite{reimers2019sbert, wolf2019huggingface}), yielding a $d_q = 768$-dimensional embedding per question to condition the participant answers~\cite{reimers2019sbert}. 


\subsubsection{Physiological features}
rPPG-derived heart rate (in beats per minute)
is estimated using the pyVHR package~\cite{boccignone2025psdclustering, boccignone2020openremoteppg, boccignone2022pyvhr} with a 6-second window and a 1-second stride, yielding one scalar estimate per window.

\subsection{Post-hoc Multi-Level Attribution via Integrated Gradients (IG)}
\label{app:interpretability}

\begin{figure*}[t]
    \centering
    \includegraphics[width=0.9\textwidth,keepaspectratio]{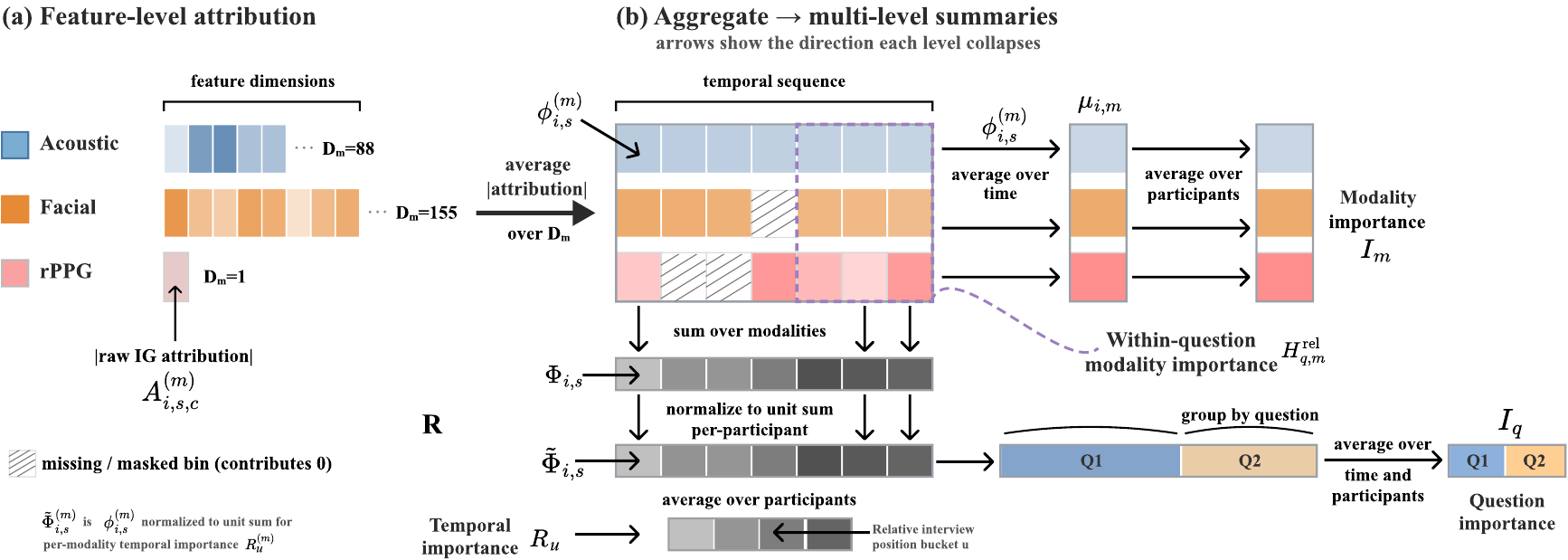}
    \caption{
    (a) After training, IG assigns a scalar attribution $A_{i,s,c}^{(m)}$ to each feature $c$ of modality-feature $m$ at temporal bin $s$ for participant $i$. Averaging the absolute attributions over the modality-feature's $D_m$ dimensions and gating by the availability mask (hatched bins are missing and contribute 0) reduces the attributions of all features $c$ to one score per temporal bin $\phi_{i,s}^{(m)}$. (b) These per-temporal-bin scores are aggregated along the different axes, and arrows indicate the dimensions along which the attributions are aggregated to produce the modality importance $I_m$, question importance $I_q$, within-question modality importance $H_{q,m}^{\mathrm{rel}}$, and temporal importance $R_u$.}
    \label{fig:attr02}
\end{figure*}

This section gives full derivations for the attribution aggregations summarized in~\autoref{sec:interpretability}. We aggregate feature-level IG~\cite{sundararajan2017integratedgradients} scores at four levels of the interview structure to produce interpretable summaries across modalities, questions, and time. The aggregation scheme across these axes is shown in~\autoref{fig:attr02}.

\subsubsection{Base attribution scores}
\label{app:base_attr}
Let $A_{i,s,c}^{(m)}$ denote the IG scalar attribution for feature $c$ of modality-feature $m$ at temporal bin $s$ for participant $i$. IG is computed over the concatenated participant-level feature time series $\mathbf{X}_i^{(m)}\in\mathbb{R}^{S_i\times D_m}$ (defined in~\autoref{sec:alignment}), so $s$ indexes temporal bins across the full interview.
Attributions are averaged across feature vector dimensionality $D_m$ within each modality-feature, so that modality-features of different dimensionalities contribute comparably to downstream aggregations~\cite{mu2026remotecognitivewellbeing}. 

This averaging produces score $\phi_{i,s}^{(m)}$, the per-temporal-bin attribution for modality-feature $m$ and participant $i$: 
\begin{equation}
\phi_{i,s}^{(m)}
= M_{i,s}^{(m)}
  \cdot \frac{1}{D_m}\sum_{c=1}^{D_m}\left|A_{i,s,c}^{(m)}\right|,
\end{equation}
where $M_{i,s}^{(m)}$ is the mask that ensures that unobserved temporal bins contribute zero attribution. Plots of $\phi_{i,s}^{(m)}$ for representative participants across their complete interviews are shown in~\autoref{fig:single_patient_profiles}.

The total attribution at temporal bin $s$ from all modality-features is $\Phi_{i,s} = \sum_{m \in \mathcal{M}} \phi_{i,s}^{(m)}$. To enable temporal dataset-level aggregation without bias from variation in interview length or modality availability across participants, attributions are normalized within each participant to sum to one:
\begin{equation}
\tilde{\Phi}_{i,s}
= \frac{\Phi_{i,s}}{\displaystyle\sum_{s'} \Phi_{i,s'}},
\qquad
\tilde{\Phi}_{i,s}^{(m)}
= \frac{\phi_{i,s}^{(m)}}{\displaystyle\sum_{s'} \Phi_{i,s'}},
\end{equation}
Here $\tilde{\Phi}_{i,s}$ is the share of participant $i$'s total attribution
that falls at temporal bin $s$, and $\tilde{\Phi}_{i,s}^{(m)}$ is the share
attributable to modality-feature $m$ at that bin. Both are expressed as fractions
of the same denominator so that $\sum_m \tilde{\Phi}_{i,s}^{(m)} = \tilde{\Phi}_{i,s}$
at every temporal bin.

\subsubsection{Modality Importance}
\label{app:mod_imp}

Let $\mathcal{M}_i = \{m : \sum_s M_{i,s}^{(m)} > 0\}$ be the set of modality-features observed at least once for participant $i$, and let $\mathcal{P} = \{i : \sum_m \sum_s M_{i,s}^{(m)} > 0\}$ be the set of all participants with at least one valid modality-feature observation.
Dividing the summed attribution for modality-feature $m$ by its number of observed bins, $\sum_s M_{i,s}^{(m)}$, gives the mean attribution per observed bin, so features present in fewer bins are not penalized for contributing fewer terms. This yields the per-modality-feature mean attribution $\mu_{i,m}$ for each participant:
\begin{equation}
\mu_{i,m}
= \frac{\displaystyle\sum_s \phi_{i,s}^{(m)}}{\displaystyle\sum_s M_{i,s}^{(m)}}.
\end{equation}
Within each participant, $\mu_{i,m}$ is normalized over observed modality-features so that contributions are comparable across participants. The per-participant normalization produces $\tilde{I}_{i,m}$:
\begin{equation}
\tilde{I}_{i,m}
= \frac{\mu_{i,m}}{\displaystyle\sum_{m' \in \mathcal{M}_i} \mu_{i,m'}}.
\end{equation}
We set $\tilde{I}_{i,m} = 0$ for participants for whom modality-feature $m$ was entirely absent. The dataset-level modality-feature importance (shown in~\autoref{fig:ig_main}(a)) aggregated over participants is:
\begin{equation}
I_m = \frac{1}{|\mathcal{P}|} \sum_{i \in \mathcal{P}} \tilde{I}_{i,m}.
\end{equation}

\subsubsection{Question Importance}
\label{app:question_importance}

Let $q_{i,s}$ denote the question identity of temporal bin $s$ for participant $i$,
let $N_{i,q} = \sum_s \mathbf{1}\{q_{i,s}=q\}$
be the number of observed bins for question $q$, and let $\mathcal{P}_q = \{i : N_{i,q} > 0\}$ be
the set of participants who answered question $q$.
We normalize $\tilde{\Phi}_{i,s}$ by $N_{i,q}$ to reduce bias toward longer answers. We then average that result over participants, producing the dataset-level question importance (shown in~\autoref{fig:ig_question_q}):
\begin{equation}
I_q
= \frac{1}{|\mathcal{P}_q|}
\sum_{i \in \mathcal{P}_q}
\frac{1}{N_{i,q}}
\sum_{s:\,q_{i,s}=q} \tilde{\Phi}_{i,s}.
\end{equation}

\subsubsection{Within-Question Modality Importance}
\label{app:q-m-importance}

Let $N_{i,q}^{(m)} = \sum_s \mathbf{1}\{q_{i,s}=q\}\,M_{i,s}^{(m)}$ be
the number of observed bins for modality-feature $m$ within question $q$ for
participant $i$, and let $\mathcal{M}_{i,q} = \{m : N_{i,q}^{(m)} > 0\}$
be the modality-features observed at least once during that question.
As described in~\autoref{app:mod_imp}, the per-bin
attribution is normalized by observed bin count to avoid penalizing
modality-features with more missing temporal bins:
\begin{equation}
\mu_{i,q,m}
= \frac{\displaystyle\sum_{s:\,q_{i,s}=q} \phi_{i,s}^{(m)}}{N_{i,q}^{(m)}}.
\end{equation}
Within each participant and question, $\mu_{i,q,m}$ is normalized over
observed modality-features so that contributions are comparable across
participants:
\begin{equation}
r_{i,q,m}
= \frac{\mu_{i,q,m}}{\displaystyle\sum_{m' \in \mathcal{M}_{i,q}} \mu_{i,q,m'}}.
\end{equation}
The dataset-level within-question modality-feature importance (shown in~\autoref{fig:ig_question_m}) is:
\begin{equation}
H_{q,m}^{\mathrm{rel}}
= \frac{1}{|\mathcal{P}_q|}
\sum_{i \in \mathcal{P}_q} r_{i,q,m}.
\end{equation}

\subsubsection{Temporal Importance}

Bins are within-answer temporal units, so their total count varies with interview length, and a bin does not correspond to a fixed recording position across participants. To place every participant on a common temporal axis for a cohort-level temporal importance summary, we group each participant's bins into $U$ equal-width buckets by relative position in the interview.  Unlike bins, which measure absolute time within a single answer and vary in number per participant, buckets measure normalized position across the whole interview and are fixed at $U$ buckets per participant. Thus, bucket $u$ denotes the same relative point in the interview for all participants. We set $U=20$, so each bucket spans $5\%$ of the normalized interview duration. 

Let $\mathcal{B}_{i,u}$ denote the set of bins for participant $i$
in bucket $u$, and let $\mathcal{P}_u = \{i : |\mathcal{B}_{i,u}| > 0\}$
be the set of participants with at least one bin in bucket $u$.
Dividing each participant's temporal-importance values $\tilde{\Phi}_{i,s}$ summed over bucket $u$ by $|\mathcal{B}_{i,u}|$ aggregates them into a single mean importance value. Therefore, a participant with many bins in that bucket does not contribute more than one with fewer bins. Averaging over the $|\mathcal{P}_u|$ participants combines the means across participants with bins in bucket $u$. Therefore, $R_u$ is the mean per-bin temporal importance at relative position $u$ in the interview:
\begin{equation}
R_u
= \frac{1}{|\mathcal{P}_u|}
\sum_{i \in \mathcal{P}_u}
\frac{1}{|\mathcal{B}_{i,u}|}
\sum_{s \in \mathcal{B}_{i,u}} \tilde{\Phi}_{i,s}.
\end{equation}
Here, $\tilde{\Phi}_{i,s}$ is the per-bin temporal importance defined in~\autoref{app:base_attr}. $R_u$ across the full interview is shown in~\autoref{fig:ig_main}(b).
The modality-feature-specific temporal importance $R_u^{(m)}$ is obtained by substituting $\tilde{\Phi}_{i,s}^{(m)}$ for $\tilde{\Phi}_{i,s}$. $R_u^{(m)}$ across the full interview for eyegaze, head pose, and ComParE is likewise shown in~\autoref{fig:ig_main}(b).

\section{Experiments}
\label{app:exp}





\begin{figure*}[t]
    \centering

    \begin{minipage}[t]{0.52\textwidth}
        \centering
        \includegraphics[
            width=\linewidth,
            height=0.21\textheight,
            keepaspectratio
        ]{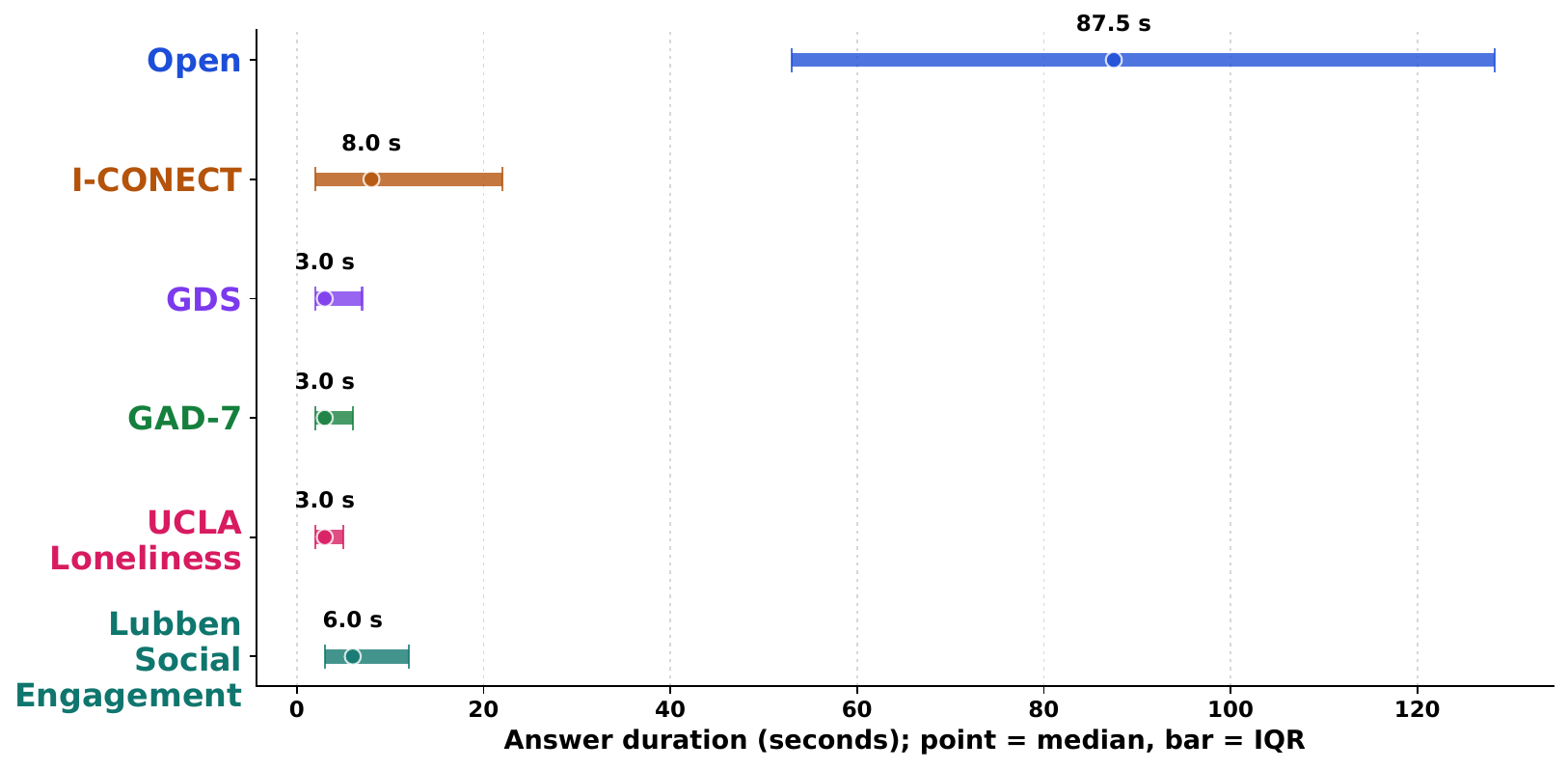}

        \vspace{0.2em}
        \textbf{(a)} Answer duration (median and IQR)
    \end{minipage}
    \hfill
    \begin{minipage}[t]{0.44\textwidth}
        \centering
        \includegraphics[
            width=\linewidth,
            height=0.19\textheight,
            keepaspectratio
        ]{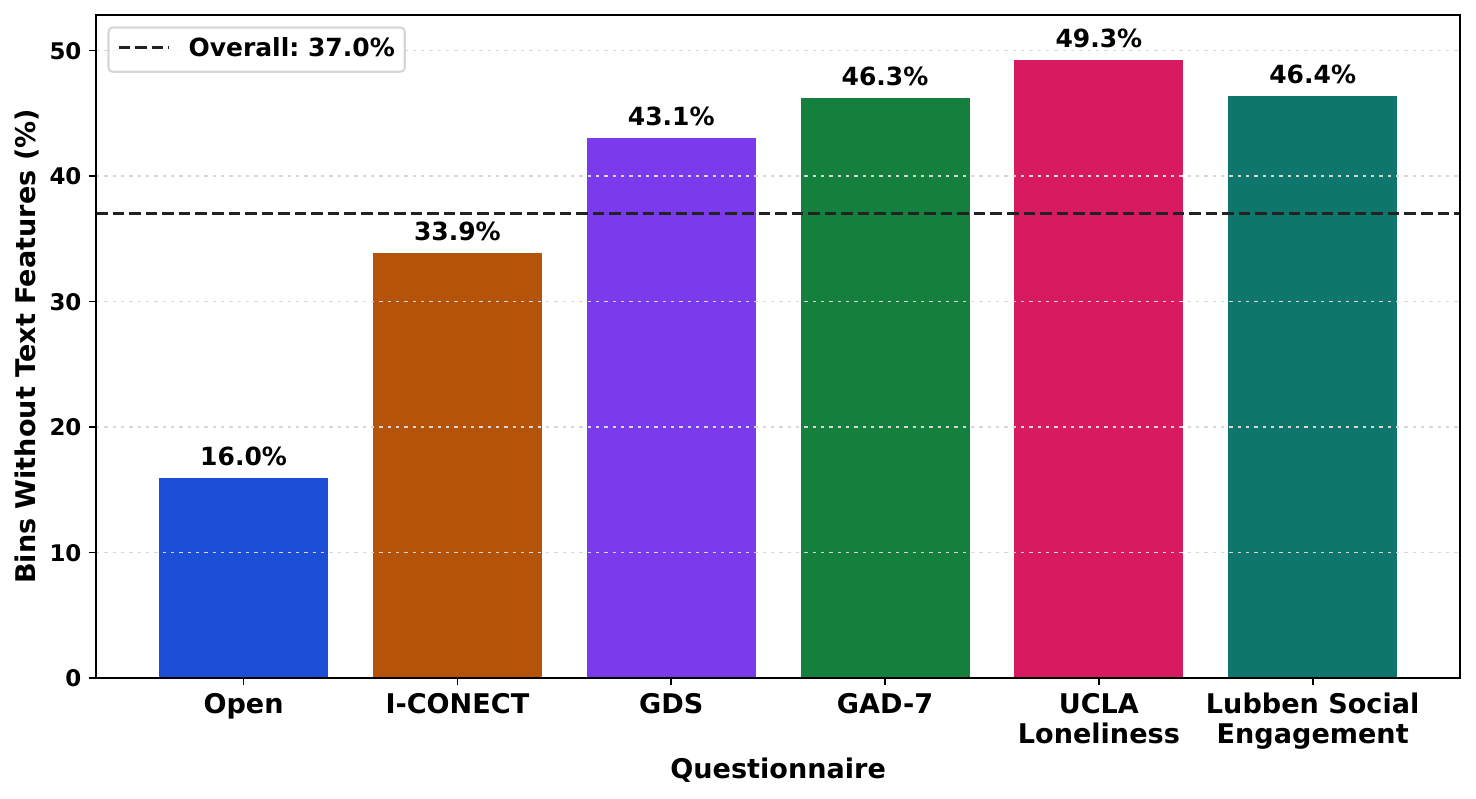}

        \vspace{0.2em}
        \textbf{(b)} Text feature bin missingness
    \end{minipage}

    \vspace{-0.5em}
    \caption{Answer duration and text-feature missingness by question group.
    (a) Median answer duration (point) and interquartile range (IQR) (bar)
    within each question group. Open-ended questions elicit longer answers than
    structured questions. (b) Percentage of answer-aligned one-second bins
    that contain no transcribed word, within each question group, with the overall
    rate (37.0\%) marked by the dashed line. Text missingness rises as answer
    duration shortens, from 16\% for open-ended questions to 49\% for the
    UCLA Loneliness questions.}
    \label{fig:txt}
\end{figure*}

\subsection{Dataset}
\label{app:dataset}

Each interview followed a fixed-order protocol comprising six consecutive question groups: four open-ended narrative prompts (including autobiographical recall and a picture description task using the Thematic Apperception Test (TAT) \cite{morgan1935tat}), an 18-item social and functional health check derived from the I-CONECT protocol \cite{yu2021iconect}, the 30-item GDS \cite{yesavage1982gds}, the GAD-7 \cite{spitzer2006gad7}, the 20-item UCLA Loneliness Scale \cite{russell1978uclaloneliness}, and the 12-item Lubben Social Network Scale (Social Engagement)~\cite{lubben1988lsns}. 


Some questions were not answered by design. The interview used skip logic, so a question could be omitted based on a prior answer. For example, participants who answered ``yes'' to whether they live alone were not asked the follow-up question of how many people live with them. Other omissions arose from changes to the administered question groups across the study, such as the change from the short to the long version of the UCLA Loneliness Scale. The TAT question was unanswered for several participants due to difficulty showing the picture over Zoom. Because the interviewer annotated answer timestamps by hand, some questions with a spoken answer were occasionally marked as unanswered.



Missingness varies across input modality-features. The acoustic features (eGeMAPS, ComParE, and Wav2Vec2) have less than 0.1\% of bins missing across all participants. Visual and physiological features have low missingness rates (action units and landmarks: 3.0\%; eyegaze: 2.6\%; head pose: 2.5\%; rPPG: 1.4\%). One participant had no valid rPPG signal across the full interview. Language has the highest per-bin missingness (37.0\%), because answer duration varies by question type. Open-ended narrative questions elicited the longest answers (median 87.5s) and had the lowest text missingness (16\%), whereas Likert-scale and yes/no questions elicited short responses (median 3--8s) and had higher missingness, including 43.1\% for the GDS, 46.3\% for the GAD-7, 49.3\% for the UCLA Loneliness scale, and 46.4\% for the Lubben Social Engagement scale (\autoref{fig:txt}). Furthermore, transcription errors are an additional source of missingness. For example, WhisperX sometimes leaves short fillers such as ``Oh'' or ``Um'' untranscribed even when spoken within an answer. In all cases, missing bins are assigned zero feature vectors and flagged by the modality-feature-timestep mask rather than excluded, as described in~\autoref{sec:missingness}.




The ComParE feature set is standardized and projected to 512 dimensions via Principal Component Analysis fitted on training-fold data, as the raw dimensionality of the feature is $6{,}373$.

Participant feature time series are formed by concatenating bins across all answers, and the resulting number of bins ranged from 597 to 2260 across participants (median 935). 


\subsection{Training and Evaluation Protocol}

To mitigate class imbalance, training batches are constructed using a class-balanced sampler that oversamples participants from the minority class. Within each fold, all models are trained for $\approx 10$ epochs. After training, each model is evaluated once on that fold’s unseen, held-out test set.

We report the AUROC score as the primary metric following previous work for mental health analysis on older adults with MCI~\cite{mu2026remotecognitivewellbeing}. We also report AUPRC, macro-F1, positive class precision, and recall scores in~\autoref{tab:extended_classification_results}. 
For significance testing, we also report pairwise statistical comparisons with one-sided Student's paired $t$-tests over fold-level AUROC scores;~\autoref{tab:appendix_pairwise_pvalues} reports whether the first model in each contrast outperformed the second.
For each task, interpretability analysis is performed on the unseen test participants for the fold corresponding to the best-performing temporally aligned ($+$T) model across all repetition-fold pairs.

\begin{figure}[t]
\centering
    \includegraphics[width=0.75\linewidth,keepaspectratio]{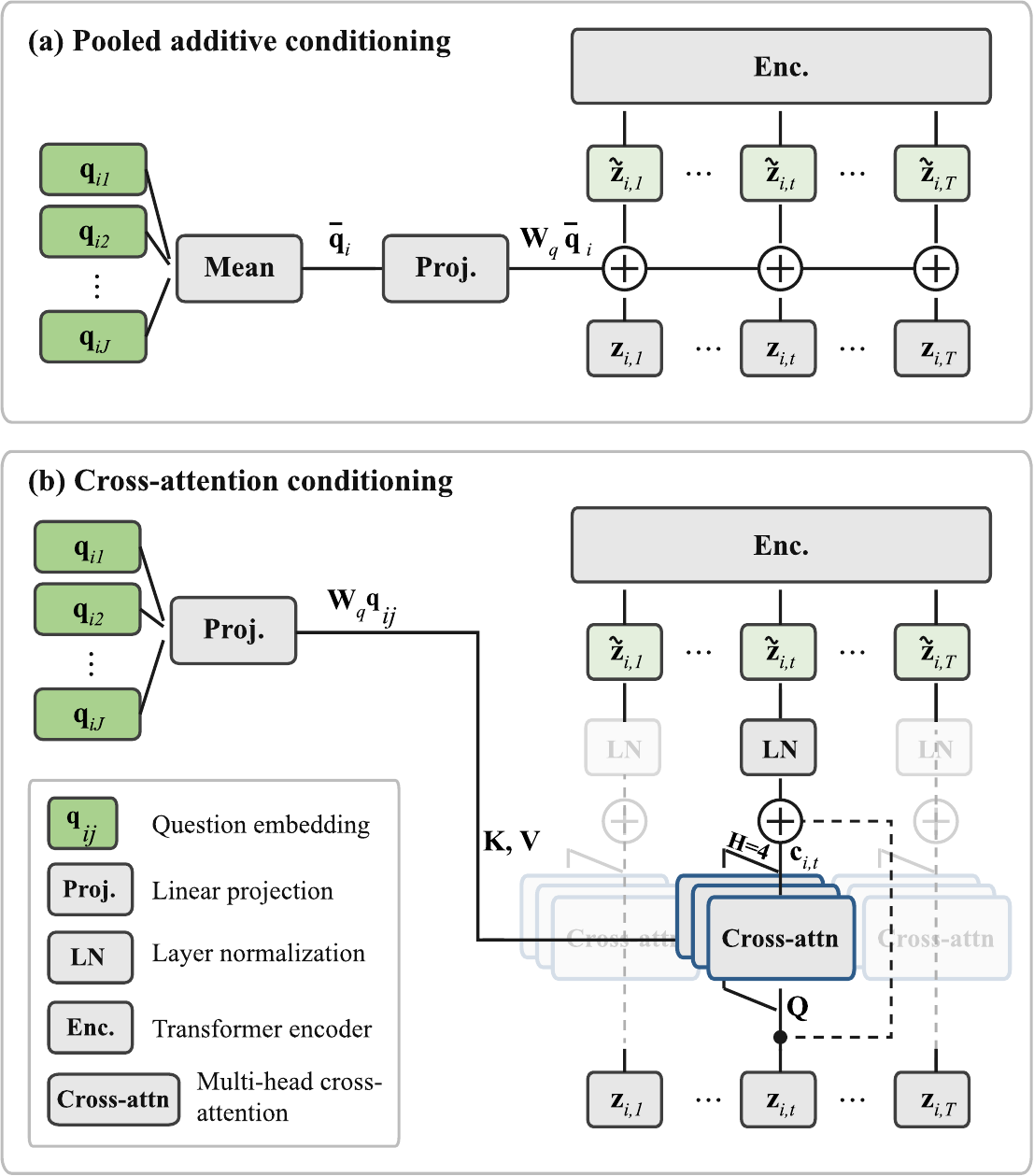}
    \caption{Question-conditioning models for the sequence-index-paired baseline. (a) Pooled additive conditioning. (b) Cross-attention conditioning. The stacked blocks denote the $H=4$ heads, and the faded blocks denote additional timesteps processed by the same attention module with the same question keys and values.
}
    \label{fig:qcond}
\end{figure}

\subsection{Ablations}
\label{app:ablations}
\subsubsection{Sequence-Index-Paired Baseline}
To evaluate the impact of temporal alignment in the ablation experiments, we create a temporally unaligned version of the dataset. 
Each modality-feature is processed as an independent
time series $\mathbf{X}^{(m)} \in \mathbb{R}^{T_m \times D_m}$, where
$T_m$ varies across modality-features according to their native sampling rates. Language features are represented as a single transcript-level embedding for the full interview rather than as a sequence of bin-level embeddings. At training time, we apply uniform
downsampling to the shortest sequence length $T$ within the batch, with $t$ indexing timesteps within a single sequence. This follows strategies similar to those used to normalize feature time series to a global common temporal length by pooling~\cite{fan2024transformermultimodaldepression} or cropping and interpolation~\cite{tao2024depmstat}. Instead, we set the target length per batch rather than globally because interview durations range from $600$ to $3{,}600$s, so a single fixed length would either discard most of the longer interviews or heavily upsample the shorter ones. This sequence-index-paired baseline follows prior approaches that pair features by their index position in the sequence rather than by timestamps during training~\cite{fan2024transformermultimodaldepression, tao2024depmstat}.

\subsubsection{Question-Conditioning Models}
\label{app:question_fusion}
\paragraph{Pooled Additive Conditioning (Linear/Non-linear + Q, sum)}
For the sequence-index-paired baseline, the multimodal feature time series are not segmented by answer timestamps or temporally aligned. We therefore adapt the additive conditioning that Zhang \textit{et al.}~\cite{zhang2025interviewerbiasdepression} apply to each question and its associated response to operate at the interview level. Their approach is discussed further in the Related Work section of the main text (\autoref{sec:related_work_qcond}). In our adaptation, all
question embeddings are averaged into a single interview-level summary $\bar{\mathbf{q}}_i$, and projected to $d$ (defined in~\autoref{sec:projection_backbones}) by $W_q$. This projected vector is then added uniformly across timesteps to the fused multimodal sequence, similar to additive global conditioning in WaveNet~\cite{oord2016wavenet}:
\begin{equation}
    \bar{\mathbf{q}}_i = \frac{1}{J_i}\sum_{j=1}^{J_i}\mathbf{q}_{ij},
    \qquad
    \tilde{\mathbf{z}}_{i,t} = \mathbf{z}_{i,t} + W_q \bar{\mathbf{q}}_i,
\end{equation}

where $W_q \in \mathbb{R}^{d\times d_q}$, $\mathbf{q}_{ij}$ is the embedding of the $j$-th question answered by participant $i$, $\mathbf{z}_{i,t}$ is the fused multimodal token at timestep $t$, and $\tilde{\mathbf{z}}_{i,t}$ is the question-conditioned multimodal token. The conditioned sequence $\tilde{\mathbf{Z}}_{i}$ formed by stacking the conditioned tokens is then passed to the Transformer encoder as described in~\autoref{sec:q_cond_pred}.

\paragraph{Cross-Attention Conditioning (Linear/Non-linear + Q, cross-attn.)}
We generalize the question-conditioned attention used in HCAG~\cite{niu2021hcagdepression} to the sequence-index-paired baseline by replacing additive attention with cross-attention. Further discussion of the HCAG approach is provided in the Related Work section of the main text (\autoref{sec:related_work_qcond}). In our adaptation, each fused multimodal token $\mathbf{z}_{i,t}$ attends to the participant's set of $J_i$ answered question embeddings through a multi-head cross-attention block~\cite{vaswani2017attention}. We use $H=4$ heads, model dimension $d=128$, per-head dimension $d_h=d/H=32$, and dropout $=0.3$. The question embeddings $\{\mathbf{q}_{ij}\}_{j=1}^{J_i}$ are projected to $d$ by $W_q$. For head $h$, the fused multimodal token $\mathbf{z}_{i,t}$ at timestep $t$ forms the query, and each projected question embedding $W_q\mathbf{q}_{ij}$ forms the keys and values:
\begin{equation}
\mathbf{Q}_{i,t}^h = W_Q^h\mathbf{z}_{i,t},\quad
\mathbf{K}_{ij}^h = W_K^h(W_q\mathbf{q}_{ij}),
\end{equation}
\begin{equation}
    \mathbf{V}_{ij}^h = W_V^h(W_q\mathbf{q}_{ij}),
\end{equation}

where $W_Q^h,W_K^h, W_V^h \in \mathbb{R}^{d_h\times d}$, and $W_q \in \mathbb{R}^{d\times d_q}$. The attention weight from timestep $t$ to question $j$, normalized across the participant's answered questions, is $\alpha_{i,t,j}^h$: 
\begin{equation}
\alpha_{i,t,j}^h =
\operatorname{softmax}_j\!\left(
\frac{(\mathbf{Q}_{i,t}^h)^\top\mathbf{K}_{ij}^h}{\sqrt{d_h}}
\right).
\end{equation}
The resulting question context $\mathbf{c}_{i,t}^h \in \mathbb{R}^{d_h}$ is the attention-weighted combination of the projected question embeddings:
\begin{equation}
\mathbf{c}_{i,t}^h = \sum_{j=1}^{J_i}\alpha_{i,t,j}^h\mathbf{V}_{ij}^h.
\end{equation}

The head question contexts are concatenated, projected by $W_O\in \mathbb{R}^{d\times d}$ to form $\mathbf{c}_{i,t}$, which is added back to the fused multimodal token $\mathbf{z}_{i,t}$ through a residual connection with dropout and layer normalization, giving  $\tilde{\mathbf{z}}_{i,t}$, the question-conditioned multimodal token:
\begin{equation}
\mathbf{c}_{i,t} = W_O\big[\mathbf{c}_{i,t}^1;\dots;\mathbf{c}_{i,t}^H\big],
\end{equation}
\begin{equation}
    \tilde{\mathbf{z}}_{i,t} =
\mathrm{LayerNorm}\!\big(\mathbf{z}_{i,t} + \mathrm{Dropout}(\mathbf{c}_{i,t})\big),
\end{equation}
where $\mathbf{c}_{i,t}, \mathbf{z}_{i,t},$ and $\tilde{\mathbf{z}}_{i,t} \in \mathbb{R}^{d}$. Each fused multimodal token therefore receives a learned mixture of all of the participant's question embeddings. The conditioned sequence is then passed to the Transformer encoder as described in~\autoref{sec:q_cond_pred}.

\section{Results}
\label{app:results}
\subsection{Overall Performance}
\label{app:metrics}

\begin{table*}[t]
\centering
\caption{
Classification results (mean $\pm$ 95\% CIs).
Precision and recall are reported for the positive class (positive depression- or anxiety-risk class)
at a classification threshold of 0.5.
\textbf{Bold}, \underline{underline}, and \textnormal{\textit{italics}}
denote the highest, second-highest, and third-highest results within each task
and metric, respectively. Tied results receive the same rank.
T denotes temporally aligned features, and Q denotes question conditioning. Linear and Non-linear denote the different projection backbones. The no-mask variant replaces the mask with an all-ones mask, effectively disabling masking, whereas the open-only variant uses only multimodal features extracted from participants’ responses to the four open-ended questions.
}
\label{tab:extended_classification_results}

\scriptsize
\setlength{\tabcolsep}{2.5pt}
\renewcommand{\arraystretch}{1.08}

\begin{tabular*}{\textwidth}{
@{\extracolsep{\fill}}lcccccccc@{}}
\toprule
& \multicolumn{4}{c}{\textbf{Depression}}
& \multicolumn{4}{c}{\textbf{Anxiety}} \\
\cmidrule(lr){2-5}
\cmidrule(lr){6-9}

\textbf{Variant}
& \textbf{AUPRC} $\uparrow$
& \textbf{Macro-F1} $\uparrow$
& \textbf{Precision}$_+$ $\uparrow$
& \textbf{Recall}$_+$ $\uparrow$
& \textbf{AUPRC} $\uparrow$
& \textbf{Macro-F1} $\uparrow$
& \textbf{Precision}$_+$ $\uparrow$
& \textbf{Recall}$_+$ $\uparrow$ \\
\midrule

\multicolumn{9}{@{}l}{\textit{Sequence-index-paired baseline} \cite{fan2024transformermultimodaldepression,tao2024depmstat}} \\

Linear
& $0.56 \pm 0.004$
& $0.45 \pm 0.08$
& $0.34 \pm 0.12$
& $0.31 \pm 0.16$
& $0.57 \pm 0.02$
& $\mathbf{0.55 \pm 0.07}$
& $\mathbf{0.49 \pm 0.15}$
& $\mathit{0.38 \pm 0.15}$ \\

Non-linear
& $0.60 \pm 0.09$
& $0.52 \pm 0.08$
& $0.50 \pm 0.18$
& $0.43 \pm 0.14$
& $\mathit{0.58 \pm 0.05}$
& $0.50 \pm 0.07$
& $0.40 \pm 0.14$
& $0.27 \pm 0.09$ \\

\midrule
\multicolumn{9}{@{}l}{\textit{Question-conditioned}} \\

Linear + Q, sum~\cite{zhang2025interviewerbiasdepression}
& $0.56 \pm 0.01$
& $0.48 \pm 0.04$
& $0.42 \pm 0.06$
& $0.43 \pm 0.13$
& $0.54 \pm 0.06$
& $0.48 \pm 0.03$
& $0.30 \pm 0.11$
& $0.26 \pm 0.07$ \\

Linear + Q, cross-attn.~\cite{niu2021hcagdepression}
& $0.50 \pm 0.06$
& $0.47 \pm 0.07$
& $0.43 \pm 0.11$
& $0.39 \pm 0.10$
& $0.54 \pm 0.03$
& $\underline{0.53 \pm 0.07}$
& $\mathit{0.43 \pm 0.19}$
& $\mathit{0.38 \pm 0.13}$ \\

Non-linear + Q, sum~\cite{zhang2025interviewerbiasdepression}
& $0.61 \pm 0.03$
& $0.51 \pm 0.08$
& $0.50 \pm 0.07$
& $0.39 \pm 0.12$
& $0.51 \pm 0.08$
& $0.50 \pm 0.07$
& $0.34 \pm 0.08$
& $\mathit{0.38 \pm 0.07}$ \\

Non-linear + Q, cross-attn.~\cite{niu2021hcagdepression}
& $0.61 \pm 0.13$
& $0.52 \pm 0.07$
& $0.46 \pm 0.12$
& $0.41 \pm 0.06$
& $0.55 \pm 0.08$
& $0.49 \pm 0.11$
& $0.33 \pm 0.22$
& $\mathit{0.38 \pm 0.26}$ \\

\midrule
\multicolumn{9}{@{}l}{\textit{Temporally aligned}} \\

Linear + T (no mask)
& $\underline{0.68 \pm 0.09}$
& $\underline{0.57 \pm 0.03}$
& $\mathbf{0.59 \pm 0.07}$
& $\underline{0.54 \pm 0.16}$
& $0.55 \pm 0.05$
& $0.49 \pm 0.07$
& $0.31 \pm 0.08$
& $0.34 \pm 0.10$ \\

Linear + T
& $\mathit{0.67 \pm 0.08}$
& $0.54 \pm 0.07$
& $\underline{0.58 \pm 0.19}$
& $\mathit{0.53 \pm 0.12}$
& $\mathbf{0.63 \pm 0.12}$
& $0.51 \pm 0.09$
& $\underline{0.48 \pm 0.29}$
& $0.34 \pm 0.18$ \\

Non-linear + T (no mask)
& $0.66 \pm 0.04$
& $0.53 \pm 0.05$
& $0.50 \pm 0.13$
& $0.36 \pm 0.10$
& $0.54 \pm 0.09$
& $0.48 \pm 0.02$
& $0.30 \pm 0.03$
& $\mathit{0.38 \pm 0.05}$ \\

Non-linear + T
& $\underline{0.68 \pm 0.04}$
& $\mathit{0.56 \pm 0.02}$
& $0.54 \pm 0.03$
& $\mathit{0.53 \pm 0.05}$
& $0.57 \pm 0.08$
& $0.47 \pm 0.05$
& $0.40 \pm 0.02$
& $0.37 \pm 0.05$ \\

\midrule
\multicolumn{9}{@{}l}{\textit{Aligned + question}} \\

Linear + Q + T
& $\mathbf{0.69 \pm 0.11}$
& $0.52 \pm 0.07$
& $\mathit{0.55 \pm 0.17}$
& $0.48 \pm 0.13$
& $\underline{0.60 \pm 0.06}$
& $\mathit{0.52 \pm 0.02}$
& $0.39 \pm 0.19$
& $\underline{0.40 \pm 0.04}$ \\

Non-linear + Q + T
& $\underline{0.68 \pm 0.06}$
& $\mathbf{0.58 \pm 0.08}$
& $0.52 \pm 0.13$
& $0.51 \pm 0.16$
& $\mathit{0.58 \pm 0.01}$
& $0.48 \pm 0.05$
& $0.34 \pm 0.14$
& $\mathbf{0.41 \pm 0.27}$ \\

\midrule
\multicolumn{9}{@{}l}{\textit{Open-ended only}} \\

Linear + T (open only)
& \multicolumn{4}{c}{--}
& $0.52 \pm 0.07$
& $0.49 \pm 0.12$
& $0.30 \pm 0.21$
& $0.37 \pm 0.19$ \\

Non-linear + T (open only)
& $\mathbf{0.69 \pm 0.08}$
& $\underline{0.57 \pm 0.06}$
& $0.50 \pm 0.05$
& $\mathbf{0.55 \pm 0.08}$
& \multicolumn{4}{c}{--} \\

\bottomrule
\end{tabular*}
\end{table*}



\begin{table}[t]
\centering
\caption{
Pairwise statistical comparisons for depression and anxiety classification.
$\Delta$ denotes the mean AUROC difference between the first and second
model. Comparisons use one-sided paired $t$-tests with the alternative
$\Delta > 0$. Bold $p$ values indicate $p < 0.05$.
}
\label{tab:appendix_pairwise_pvalues}

\scriptsize
\renewcommand{\arraystretch}{1.05}

\resizebox{\columnwidth}{!}{%
\begin{tabular}{lrrrr}
\toprule
& \multicolumn{2}{c}{\textbf{Depression}}
& \multicolumn{2}{c}{\textbf{Anxiety}} \\
\cmidrule(lr){2-3}
\cmidrule(lr){4-5}

\textbf{Contrast}
& $\boldsymbol{\Delta}$ $\uparrow$
& $\boldsymbol{p}$
& $\boldsymbol{\Delta}$ $\uparrow$
& $\boldsymbol{p}$ \\
\midrule

\multicolumn{5}{@{}l}{\textit{Linear}} \\

Q (sum) $>$ Baseline
& $\phantom{-}0.01$ & $0.31$
& $-0.03$ & $0.80$ \\

Q (cross-attn.) $>$ Baseline
& $-0.06$ & $0.96$
& $-0.02$ & $0.77$ \\

T $>$ Baseline
& $\phantom{-}0.13$ & $0.05$
& $\phantom{-}0.11$ & $0.05$ \\

T (mask) $>$ T (no mask)
& $-0.009$ & $0.56$
& $\phantom{-}0.04$ & $0.12$ \\

T (all questions) $>$ T (open only)
& \multicolumn{2}{c}{--}
& $\phantom{-}0.09$ & $\mathbf{0.02}$ \\

Q+T $>$ Baseline
& $\phantom{-}0.16$ & $\mathbf{0.009}$
& $\phantom{-}0.09$ & $0.09$ \\

Q+T $>$ Q (sum)
& $\phantom{-}0.15$ & $\mathbf{0.009}$
& $\phantom{-}0.12$ & $\mathbf{0.04}$ \\

Q+T $>$ Q (cross-attn.)
& $\phantom{-}0.22$ & $\mathbf{0.001}$
& $\phantom{-}0.12$ & $0.06$ \\

Q+T $>$ T
& $\phantom{-}0.03$ & $0.28$
& $-0.02$ & $0.64$ \\

\midrule
\multicolumn{5}{@{}l}{\textit{Non-linear}} \\

Q (sum) $>$ Baseline
& $-0.03$ & $0.73$
& $-0.02$ & $0.66$ \\

Q (cross-attn.) $>$ Baseline
& $-0.02$ & $0.62$
& $\phantom{-}0.000$ & $0.50$ \\

T $>$ Baseline
& $\phantom{-}0.10$ & $0.06$
& $\phantom{-}0.02$ & $0.39$ \\

T (mask) $>$ T (no mask)
& $\phantom{-}0.02$ & $0.34$
& $\phantom{-}0.02$ & $0.32$ \\

T (all questions) $>$ T (open only)
& $\phantom{-}0.01$ & $0.43$
& \multicolumn{2}{c}{--} \\

Q+T $>$ Baseline
& $\phantom{-}0.09$ & $0.08$
& $\phantom{-}0.05$ & $0.21$ \\

Q+T $>$ Q (sum)
& $\phantom{-}0.13$ & $\mathbf{0.02}$
& $\phantom{-}0.07$ & $0.10$ \\

Q+T $>$ Q (cross-attn.)
& $\phantom{-}0.11$ & $\mathbf{0.001}$
& $\phantom{-}0.05$ & $0.18$ \\

Q+T $>$ T
& $-0.009$ & $0.60$
& $\phantom{-}0.03$ & $0.20$ \\

\bottomrule
\end{tabular}%
}

{\scriptsize
\raggedright
T = temporally aligned features; Q = question conditioning; Q+T =
question conditioning with temporal alignment; Baseline =
sequence-index-paired model. Mask contrasts compare the real temporal
mask against an all-ones mask. Open-only models use only multimodal
features from responses to the four open-ended questions. Each comparison
uses 15 paired observations ($df=14$).
\par
}
\end{table}

\subsection{Question Importance}
\autoref{fig:ig_question_q} decomposes the attribution rankings by question groups for both tasks. For depression (\autoref{fig:ig_question_q}(a)), the highest-attribution questions center on social contact and support, self-rated health, and open-ended self-description. The top five questions are Q10 (I-CONECT: overnight visitors), Q89 (Social Engagement: availability of a relative for decision support), Q5 (I-CONECT: self-rated health), Q1 (open-ended: ``Tell me about yourself''), and Q90 (Social Engagement: providing decision support to a friend), with attributions of $(2.151, 1.702, 1.681, 1.648,\text{ and } 1.509)\times 10^{-3}$, respectively. 
For anxiety (\autoref{fig:ig_question_q}(b)), the top questions span a narrower range of $0.24 \times 10^{-3}$ and center on crying and social contact with friends. Q47 (GDS: crying frequency) leads at $1.759 \times 10^{-3}$, followed by Q91 (Social Engagement: availability of a friend for decision support, $1.636\times 10^{-3}$), Q90 (Social Engagement: providing decision support to a friend, $1.578\times 10^{-3}$), Q20 (I-CONECT: written communication with friends or family, $1.578\times 10^{-3}$), and Q17 (I-CONECT: contact with friends, $1.522 \times 10^{-3}$). 

\paragraph{Within-question modality-feature importance}
\autoref{fig:ig_question_m} shows the within-question modality-feature share ($H_{q,m}^\mathrm{rel}$) for the top-ranked questions in each task. The modality-feature rankings from~\autoref{fig:ig_main}(a) are preserved at the question level. For depression, eyegaze remains the top contributor across every shown question, followed by ComParE. For anxiety, eyegaze and head pose retain their order of importance regardless of question group.

\subsection{Single Participant Temporal Importance}
\label{app:single_temporal_importance}
\begin{figure*}[!t]
\centering
\captionsetup{
    font=footnotesize,
    skip=2pt
}

\begin{minipage}[t][0.92\textheight][t]{\textwidth}
\centering

\begin{minipage}[t]{0.49\textwidth}
    \centering
    \includegraphics[
        width=\linewidth,
        height=0.40\textheight,
        keepaspectratio,
        trim=0.2cm 0.2cm 0.07cm 0.2cm,
        clip
    ]{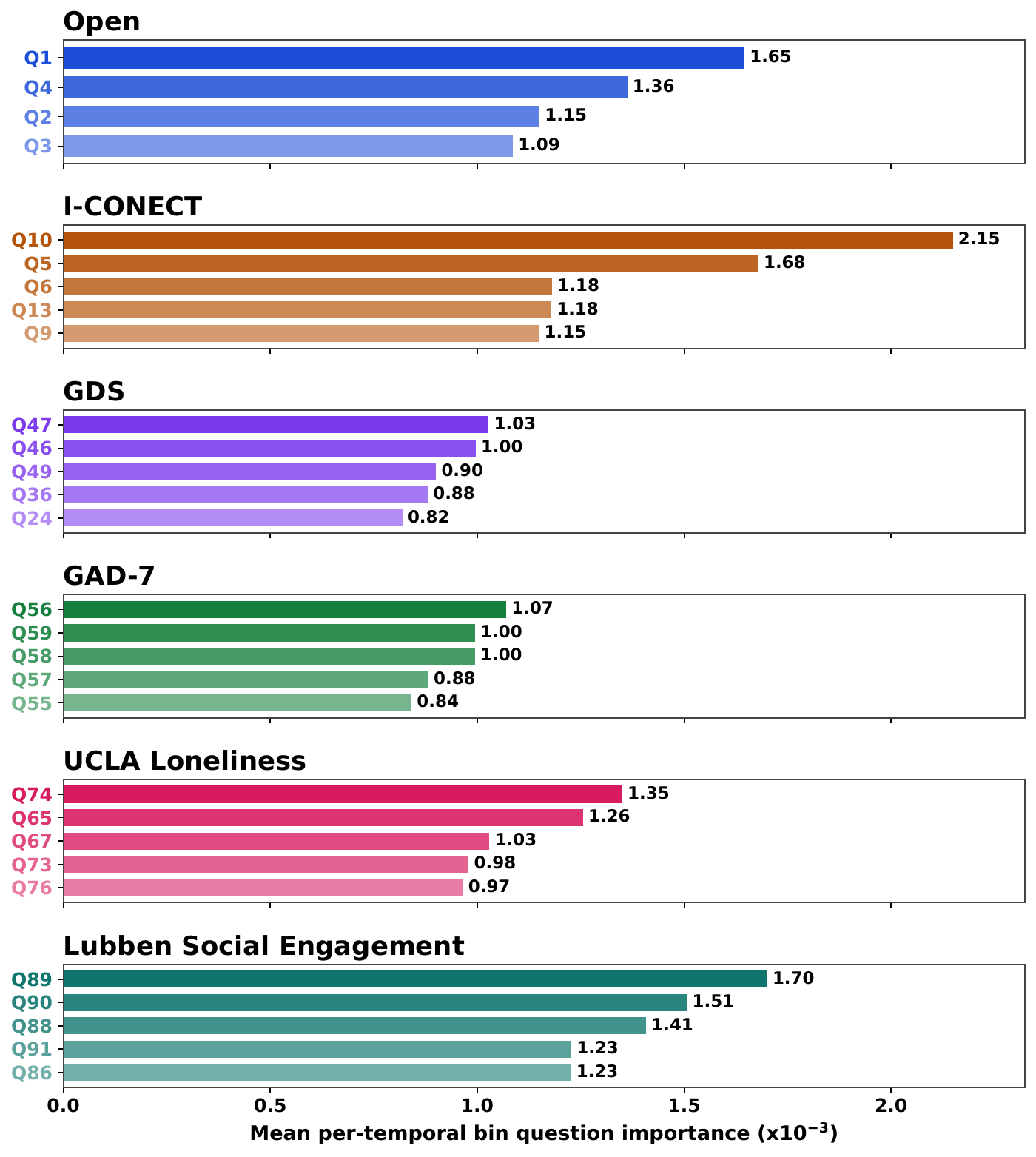}

    \textbf{(a)} Depression
\end{minipage}
\hfill
\begin{minipage}[t]{0.49\textwidth}
    \centering
    \includegraphics[
        width=\linewidth,
        height=0.40\textheight,
        keepaspectratio,
        trim=0.2cm 0.2cm 0.07cm 0.2cm,
        clip
    ]{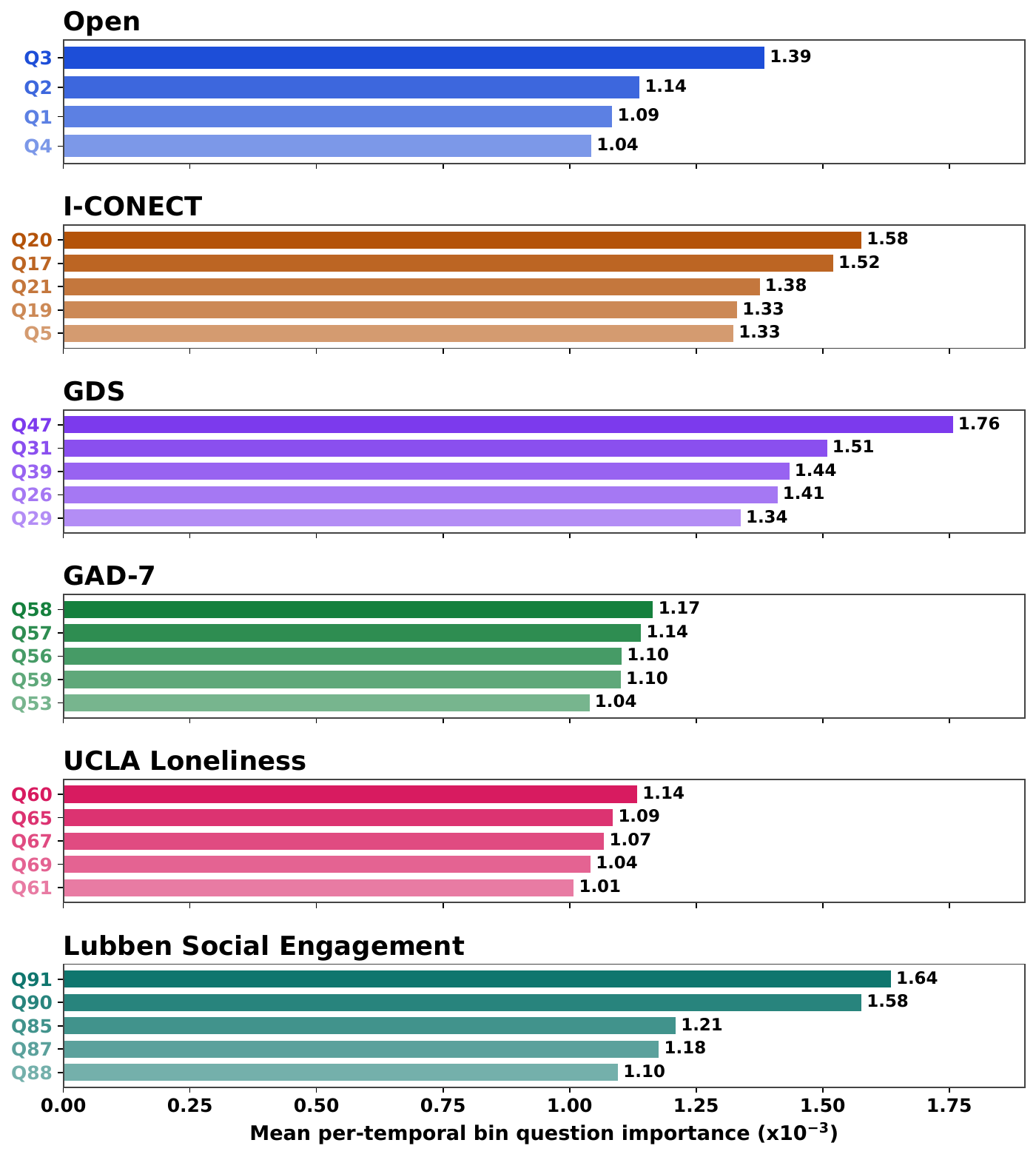}

    \textbf{(b)} Anxiety
\end{minipage}

\caption{IG question-level importance grouped by question group for depression and
anxiety prediction. Bars show normalized question-attribution scores for each
question group.}
\label{fig:ig_question_q}

\vfill

\begin{minipage}[t]{0.49\textwidth}
    \centering
    \includegraphics[
        width=\linewidth,
        trim=0.3cm 0.3cm 0.1cm 0.08cm,
        clip
    ]{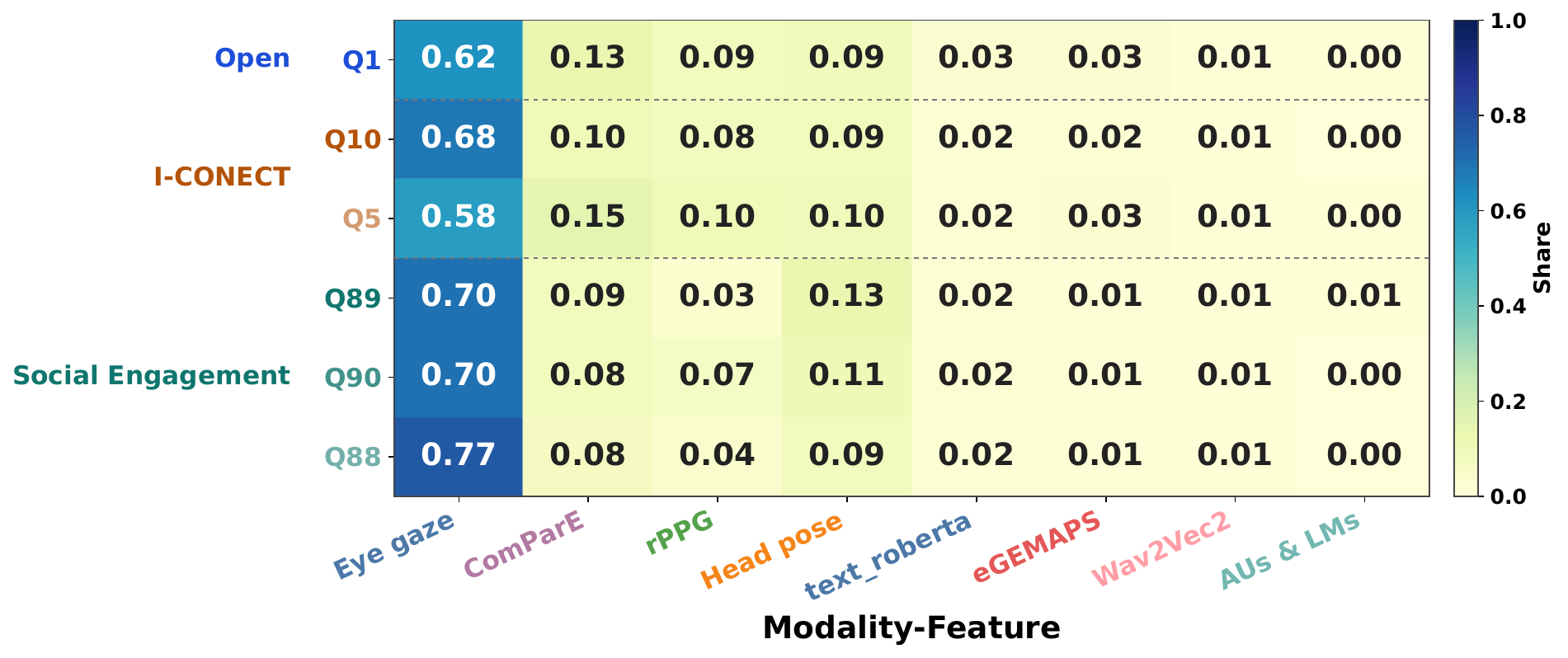}

    \textbf{(a)} Depression
\end{minipage}
\hfill
\begin{minipage}[t]{0.49\textwidth}
    \centering
    \includegraphics[
        width=\linewidth,
        trim=0.3cm 0.3cm 0.1cm 0.08cm,
        clip
    ]{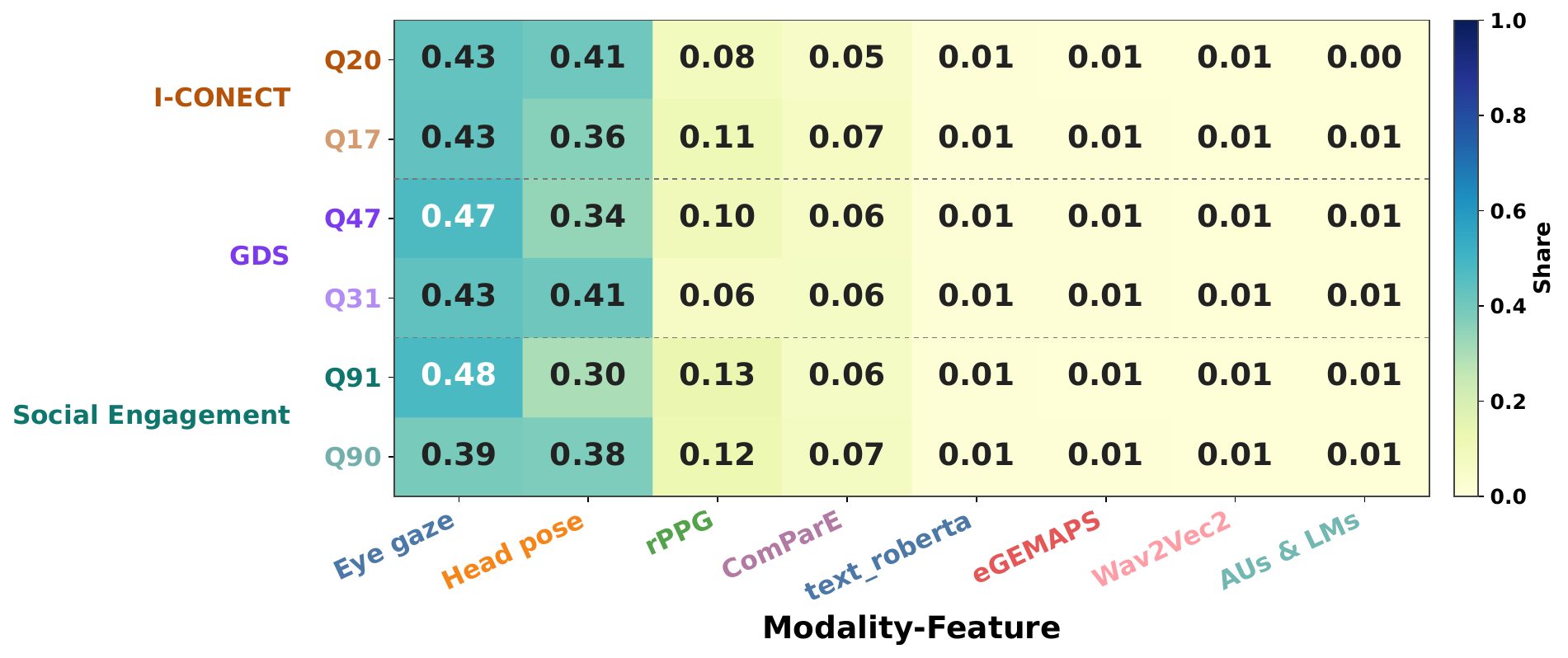}

    \textbf{(b)} Anxiety
\end{minipage}

\caption{IG within-question modality-feature importance for depression and
anxiety prediction. Each row shows the normalized modality-feature-attribution share
for a high-ranking interview question. Across question groups, eyegaze remains dominant for depression, whereas anxiety shows a more balanced attribution between eyegaze and head pose.}
\label{fig:ig_question_m}

\vfill

\begin{minipage}[t]{0.49\textwidth}
    \centering
    \includegraphics[
        width=\linewidth,
        trim=0cm 0cm 0cm 0.9cm,
        clip
    ]{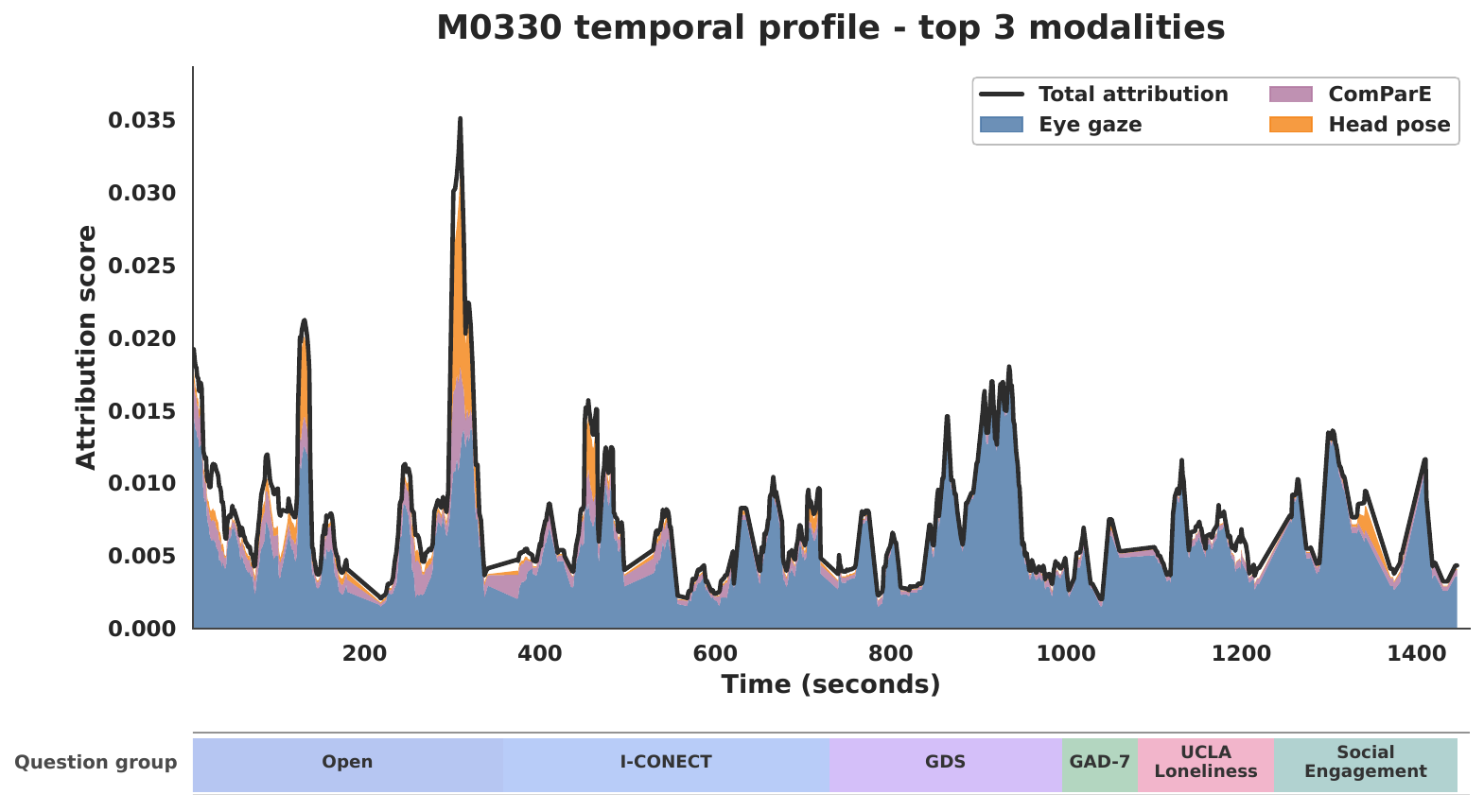}

    \textbf{(a)} Depressed participant
\end{minipage}
\hfill
\begin{minipage}[t]{0.49\textwidth}
    \centering
    \includegraphics[
        width=\linewidth,
        trim=0cm 0cm 0cm 0.9cm,
        clip
    ]{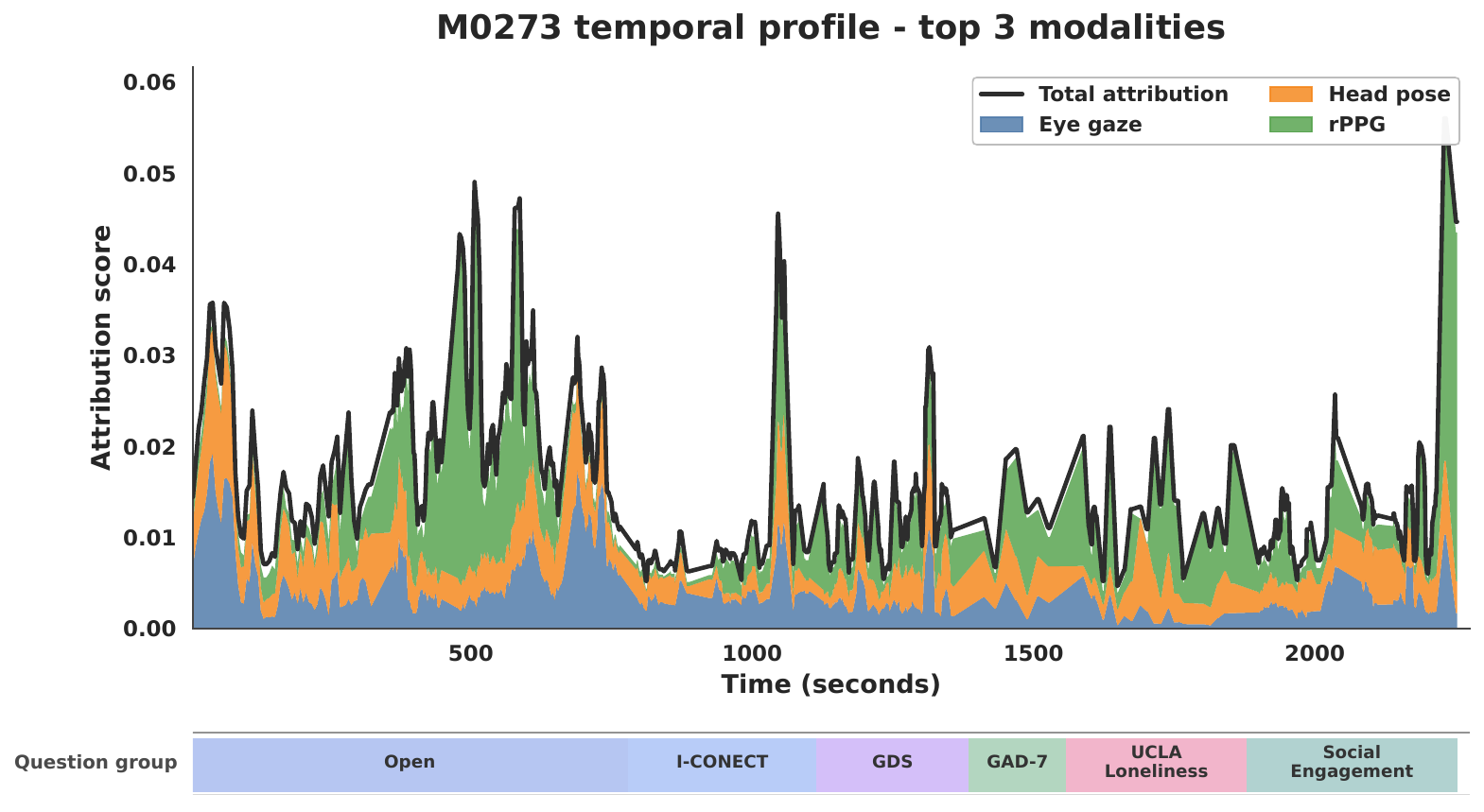}

    \textbf{(b)} Anxious participant
\end{minipage}

\caption{Single-participant IG temporal importance for the three highest-attribution modality-features. For depression, these features are eyegaze, ComParE, and head pose. Eyegaze dominates throughout the interview, with peaks during the open-ended question block. For anxiety, the top features are eyegaze, head pose, and rPPG. Eyegaze and head pose contribute comparably across all questions, while rPPG shows a peak in the open-ended question block. The colored
strip below each panel marks question-group boundaries.}
\label{fig:single_patient_profiles}

\end{minipage}
\end{figure*}

\autoref{fig:single_patient_profiles} shows IG attribution over interview time for one representative participant per task, for the three highest-contributing modality-features from~\autoref{fig:ig_main}(a). Each participant comes from the test fold of the best-performing model for the corresponding task (Non-linear+T for depression and Linear+T for anxiety). For each task, we selected a repetition–fold combination with a test AUROC $>0.8$ and chose the most confidently classified positive case from that fold (confidence $>0.9$). The attribution scores were smoothed using a 15-s window for plotting.

For the depression participant (\autoref{fig:single_patient_profiles}(a)), eyegaze has the highest attribution across all questions. Total attribution peaks during the open-ended block, driven by a sharp increase in eyegaze. ComParE contributes mainly at the beginning of the interview. Attribution decreases through the middle of the interview. This pattern matches the temporal importance findings (\autoref{fig:ig_main}(b)), where eyegaze drives the early-interview concentration for depression.

For the anxiety participant (\autoref{fig:single_patient_profiles}(b)), eyegaze and head pose remain the main contributors across all questions with comparable attribution. rPPG contributes mainly during the open-ended block and decreases afterward. This pattern matches the population-level finding that anxiety attribution is distributed more uniformly across question groups, with comparable contributions from eyegaze and head pose and without the early attribution concentration observed for depression (\autoref{fig:ig_main}(b)).

\makeatletter
\setlength{\@dblfptop}{0pt}
\setlength{\@dblfpbot}{0pt plus 1fil}
\makeatother

\clearpage

\end{document}